%% file: main.tex
\documentclass[10pt]{article} 
\usepackage[preprint]{tmlr}

\input{math_commands.tex}

\usepackage{hyperref}
\usepackage{url}
\usepackage{graphicx}
\usepackage{subcaption}
\usepackage{wrapfig}
\usepackage{adjustbox}
\usepackage{multirow}
\usepackage{booktabs}
\usepackage{float}
\usepackage{algorithmic}
\usepackage{algorithm}
\usepackage{amssymb}

\title{Continual Policy Consolidation for Lifelong Robot Learning}

\author{\name Qijun He \email u3014062@connect.hku.hk \\
      \addr Department of Data and Systems Engineering \\
      The University of Hong Kong 
      \AND
      \name Yuxuan Li \email yli289@connect.hkust-gz.edu.cn \\
      \addr Department of Computer Science\\
      Zhejiang University
      \AND
      \name Mingqi Yuan \\
      \addr The Hong Kong Polytechnic University
      \AND
      \name Xiaoquan Sun \\
      \addr Huazhong University of Science and Technology
      \AND
      \name Wentse Chen \\
      \addr School of Computer Science, Carnegie Mellon University
      \AND
      \name Jeff Schneider \\
      \addr School of Computer Science, Carnegie Mellon University
      \AND
      \name Jiayu Chen \\
      \addr The University of Hong Kong 
}

\def\month{MM}  
\def\year{YYYY} 
\def\openreview{\url{https://openreview.net/forum?id=XXXX}} 

\begin{document}

\maketitle

\begin{abstract}
Building a generalist robot policy requires continuously integrating new skills while preserving previously acquired behaviors. Directly optimizing a single policy over a growing task stream is difficult because robotic interaction is expensive, task distributions are heterogeneous, and sequential updates induce interference. To address these problems, we propose continual policy consolidation (CPC), a teacher--student framework that combines continual policy distillation with prioritized experience replay and expandable experts. This architecture separates skill acquisition from policy consolidation: teachers are trained independently through reinforcement learning, and their behaviors are continually distilled into a central generalist student. This decomposition retains the practical strength of reinforcement learning for task-specialized training while casting student-side consolidation as a supervised policy-learning problem. To balance stability and plasticity as the task stream grows, the student combines an expandable Transformer-based mixture-of-experts architecture with prioritized trajectory replay. Extensive experiments show that the student recovers a large proportion of teacher performance while achieving near-zero forgetting. These results demonstrate a scalable route for consolidating independently acquired robot skills into a continually growing generalist policy.
\end{abstract}

\section{Introduction}
\label{sec:introduction}

Robot learning is moving beyond the conventional one-policy-per-task paradigm toward generalist policies that reuse heterogeneous experience across tasks and embodiments~\citep{reed2022generalist,openx2024embodiment}. Extending such policies over a robot's lifetime gives rise to continual robot learning, where new skills must be acquired without erasing previously learned behaviors~\citep{liu2023libero,Wolczyk2021continualworldroboticbenchmark}. Directly updating a single generalist policy on a growing task stream is difficult: repeated environment interaction is costly, heterogeneous task distributions induce optimization interference, and sequential updates create a tension between plasticity for new skills and stability for existing ones~\citep{dohare2023maintaining,hendawy2023multi}. A practical alternative is to train task-specialized experts independently -- and therefore potentially in parallel -- and consolidate their behaviors into a shared student through policy distillation~\citep{rusu2015policy,traore2019discorl}. As new experts continue to arrive, however, this aggregation becomes a continual rather than one-shot distillation problem: the student must absorb new behaviors without revisiting all former teachers or erasing previously consolidated skills.

A straightforward strategy is to retain all historical demonstrations and jointly re-distill the student on the union of previous and current tasks whenever new experts arrive, following the multi-teacher consolidation setting of policy distillation~\citep{rusu2015policy}. Although joint re-distillation directly exposes the student to all earlier behaviors, both storage and repeated optimization grow with the task stream. Continual-learning methods reduce this dependence on full historical access in two main ways. Replay-based approaches preserve a representative subset of past experience for rehearsal~\citep{isele2018selectiveexperiencereplaylifelong,rolnick2019experiencereplaycontinuallearning}, whereas regularization-based approaches constrain updates to parameters that are important for earlier tasks, as exemplified by elastic weight consolidation (EWC)~\citep{kirkpatrick2017overcoming}. These strategies nevertheless introduce complementary limitations: a bounded buffer may omit rare but important behaviors, while overly restrictive parameter protection can reduce plasticity. Moreover, existing continual policy-distillation studies commonly evaluate relatively short task sequences~\citep{traore2019discorl,li2024continual}, leaving open whether a single student can consolidate a long stream of independently trained robotic experts from only a small, representative memory. This motivates our central question: how can a generalist student retain previously distilled skills while continuing to absorb new experts with limited historical demonstrations and adaptable student capacity?

\begin{figure}[t!]
    \begin{center}
    \includegraphics[width=\linewidth]{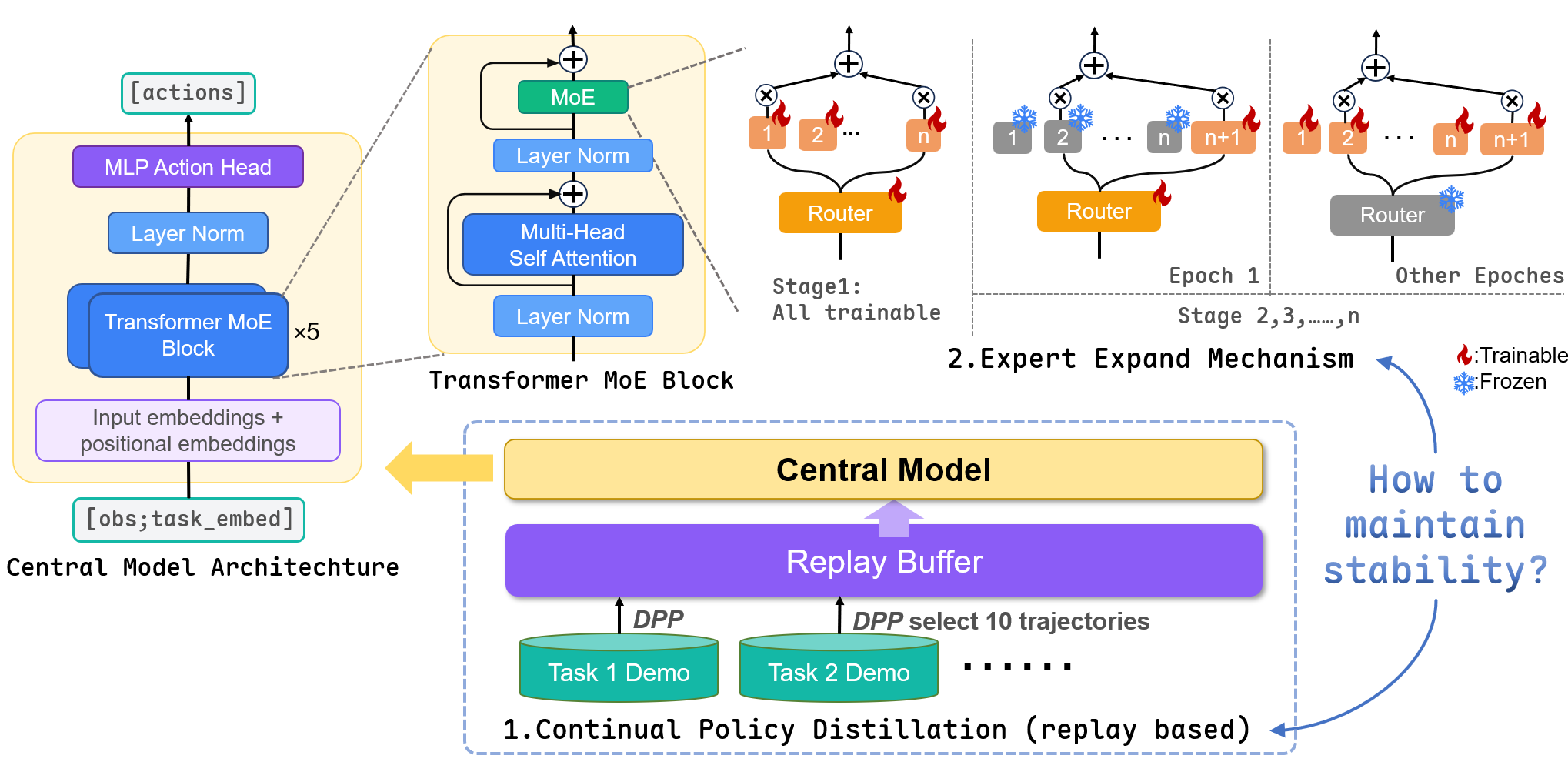}
    \end{center}
    \caption{Two complementary mechanisms for continual policy consolidation: (1) prioritized trajectory replay retains less than 10\% of the demonstration data through DPP-based selection, providing historical supervision for stability; (2) expert expansion adds capacity for plasticity, while parameter-wise freezing limits interference with previously consolidated behaviors.}
    \label{fig:fw1}
\end{figure}

In this paper, we propose continual policy consolidation (CPC), a teacher--student framework that separates robot skill acquisition from long-term skill integration. Distributed reinforcement-learning workers independently acquire task-specialized policies, while a central student continually distills their behaviors into a single generalist policy, as illustrated in Figure~\ref{fig:fw1}. Our contributions are threefold. First, we formulate a scalable consolidation process that converts sequential expert integration into continual supervised policy learning without requiring simultaneous access to all teachers. Second, we use prioritized trajectory replay to preserve a compact but behaviorally diverse memory of earlier skills and train a context encoder to provide task-discriminative inputs. Third, we develop an expandable Transformer-MoE student with a parameter-wise freezing mechanism, providing additional capacity for new tasks while protecting previously consolidated knowledge. Evaluations on Meta-World and Continual World show that the resulting framework maintains a favorable plasticity--stability balance over both short and long task streams. Supplementary LIBERO experiments further examine the replay--expansion design in a different robot-policy architecture.

\section{Related Work}

\subsection{Plasticity-Stability Dilemma in Continual Learning}
\label{sec:dilemma_cl}

Continual learning requires a model to acquire new tasks without erasing previously learned behavior. Replay preserves old supervision through stored samples~\citep{rolnick2019experiencereplaycontinuallearning,isele2018selectiveexperiencereplaylifelong}, while regularization and gradient constraints restrict updates that would damage prior knowledge~\citep{kirkpatrick2017overcoming,chaudhry2018efficient}. These approaches improve stability, but their effectiveness depends on how well a finite replay set covers earlier behaviors and how much parameter freedom remains after repeated constraints. Redundant replay consumes memory and computation by repeatedly revisiting similar examples, whereas increasingly restrictive updates can gradually reduce plasticity~\citep{dohare2023maintaining}.

Architectural methods address the complementary capacity problem. Progressive networks allocate task-specific parameters~\citep{rusu2016progressiveneuralnetworks}, and sparsely gated mixture-of-experts (MoE) models route inputs through specialized computation~\citep{shazeer2017outrageously,fedus2022switch}. Expandable expert architectures can further add capacity as the data distribution changes~\citep{chen2023lifelonglanguagepretrainingdistributionspecialized}. Expansion alone, however, does not identify which old behaviors must be preserved; updates to shared layers or routers can still change previous predictions. Stability-oriented supervision and plasticity-oriented capacity are therefore complementary. CPC coordinates them through prioritized trajectory replay and expandable experts in a single central policy.

\subsection{Continual Robot Learning}
\label{sec:related_cpd}

Continual robot learning (CRL) studies sequential skill acquisition under the stability--plasticity tension~\citep{khetarpal2022towards}, spanning online RL in Continual World and demonstration-driven learning in LIBERO~\citep{Wolczyk2021continualworldroboticbenchmark,liu2023libero}. Online approaches update a policy through environment interaction, using mechanisms such as consolidation with historical policies at different timescales~\citep{kaplanis2019policyconsolidation} or recurrent task inference combined with replay-based RL~\citep{caccia2023taskagnostic}. Online interaction can correct current-policy errors and refresh old skills when earlier environments are revisited. However, new-task interaction supplies no old-task supervision, and compact replay is difficult in actor--critic CRL because sparse past transitions must stabilize both the actor and a bootstrapped critic under distribution shift~\citep{Wolczyk2021continualworldroboticbenchmark}. Skill acquisition and retention therefore remain coupled, motivating their separation.

\textbf{Continual Policy Distillation} (CPD) separates these roles: task-specialized teachers acquire skills, and a central student learns to reproduce their actions through supervised training~\citep{rusu2015policy}. DisCoRL~\citep{traore2019discorl} demonstrates sequential consolidation of independently trained teachers, while~\citet{li2024continual} show that replay can protect previously distilled robotic controllers. This decomposition leaves the teacher-side RL algorithm independent of the central consolidation procedure, but the student remains a continual learner. When former teachers and their complete demonstrations are unavailable, it must decide which behavioral evidence to retain; as the teacher policies become more heterogeneous, a fixed-capacity student may also lose the ability to absorb new skills.

A complementary line of lifelong imitation work uses parameter-efficient modules and learned routing to allocate reusable task-dependent capacity~\citep{lei2026dmpel}. These modular approaches and replay-based continual policy distillation emphasize two distinct student-side controls: historical supervision for stability and adaptable capacity for plasticity. CPC addresses this balance by coordinating both within one central distillation model: DPP-based trajectory replay preserves diverse old behaviors, expert expansion adds capacity, and parameter protection limits interference with previously learned behaviors. 


\section{Preliminaries and Problem Formulation}
\label{sec:preliminaries}

\subsection{Markov Decision Process for Robot Learning}
\label{sec:mdp_cl}

The robot learning problem for task $k$ can be formulated as a finite-horizon Markov decision process (MDP): $\mathcal{M}_k=\langle\mathcal{S},\mathcal{A},\mathcal{T}_k,r_k,\mu_0^k,H_k,\gamma\rangle$~\citep{puterman2014markov,liu2023libero}. Here, $\mathcal{S}$ and $\mathcal{A}$ are the state and action spaces, $\mathcal{T}_k$ is the transition function, $r_k$ is the reward function, $\mu_0^k$ is the initial-state distribution, $H_k$ is the episode horizon, and $\gamma$ is the discount factor used during RL training. We express the teacher's task objective in terms of expected undiscounted episode return:
\begin{equation}
    \pi_k^{\mathrm{T}}\in\arg\max_{\pi}J_k(\pi),
    \qquad
    J_k(\pi)=\mathbb{E}\left[\sum_{t=0}^{H_k-1}r_k(s_t,a_t)\right].
    \label{eq:teacher_rl_objective}
\end{equation}

\subsection{Lifelong Imitation Learning}

In lifelong robot learning, a robot sequentially learns $K$ tasks $\{T_1,T_2,\ldots,T_K\}$ using a single policy $\pi$. Tasks may differ in their initial-state distributions and reward functions while sharing state and action spaces~\citep{liu2023libero}. When learning task $k$, the objective is to perform well across all tasks encountered so far:
\begin{equation}
    \max_{\pi}J(\pi)=\frac{1}{k}\sum_{p=1}^{k}\left[ \underset{s_t^p, a_t^p \sim \pi(\cdot;T_p), \mu_0^p}{\mathbb{E}} \left[ \sum_{t=1}^{H_p} r_p(s_t^p, a_t^p) \right] \right].
\end{equation}

As mentioned in Section~\ref{sec:related_cpd}, an imitation-based approach to lifelong robot learning uses demonstration datasets collected from a sequence of task-specialized teachers. Teacher $k$ generates demonstrations $\mathcal{D}_k=\{\tau_i^k\}_{i=1}^{N}$, where $\tau_i^k=\{(o_{i,t}^k, a_{i,t}^k)\}_{t=0}^{H_i^k-1}$. Let $o_{i,t}^k$ denote the latest observation and $\mathbf{c}_k$ the task-level context representation. We perform behavioral cloning and use mean squared error (MSE) as the distillation loss, so the central student $\pi_\theta$ learns teacher actions using the following objective:
\begin{equation}
    \mathcal{L}_k(\theta)
    =
    \mathbb{E}_{\tau_i^k\sim\mathcal{D}_k}
    \left[
        \frac{1}{H_i^k}\sum_{t=0}^{H_i^k-1}
        \left\|\mu_\theta(o_{i,t}^k,\mathbf{c}_k)-a_{i,t}^k\right\|_2^2
    \right].
    \label{eq:taskwise_distillation}
\end{equation}
Here, $\mu_\theta$ is the mean action predicted by the student, and $\mathbf{c}_k$ helps it distinguish task-specific behaviors. Thus, environment interaction is used to acquire teacher skills, whereas the central student is trained by supervised imitation. When extending this objective to lifelong imitation learning, the policy should minimize the average distillation loss over the first $k$ tasks. An important feature of lifelong learning, however, is that the student loses full access to demonstrations from the previous $k-1$ tasks while learning the $k$-th task. Approximating this joint objective with limited historical data is therefore the central problem of lifelong imitation learning.

The consolidation unit is a stage, which need not correspond to a single task. Let $\mathcal{G}_s$ denote the teachers arriving at stage $s$, $\mathcal{D}_s$ their current demonstrations, and $\mathcal{B}_{s-1}$ the selected demonstrations retained from earlier stages. The student is updated on $\mathcal{D}_s\cup\mathcal{B}_{s-1}$, so the same formulation accommodates either one incoming teacher or a group of teachers without changing the learning objective. The challenge is not simply to fit each new dataset: it is to retain a useful approximation to earlier teacher behaviors after their full datasets are no longer available. Replay determines which historical supervision remains accessible, while the student's trainable capacity determines how effectively it can reconcile that supervision with new behaviors.





\section{Methodology}
\label{sec:methodology}

Continual policy consolidation (CPC) decomposes continual robot learning into distributed skill acquisition and centralized skill consolidation. Task-specialized teachers acquire skills through RL, whereas a single student consolidates their behavior through offline continual policy distillation~\citep{rusu2015policy,li2024continual}. On the student side, prioritized trajectory replay preserves representative historical supervision, and expandable Transformer-MoE experts provide capacity for a growing teacher stream. Figure~\ref{fig:fw1} illustrates the model and data flow, while Algorithm~\ref{alg:cpd} summarizes the complete stage-wise pipeline. The following subsections specify the teacher--student construction, the replay and context mechanisms, and the expansion and protection of the expert library.

\begin{algorithm}[t]
    \vspace{0em}
    \caption{Continual distillation in CPC}
    \label{alg:cpd}
    \begingroup
    \let\AND\relax
    \begin{algorithmic}[1]
        \REQUIRE trained task/teacher groups $\{\mathcal{G}_s\}_{s=1}^{S}$; student $\pi_\theta$; context encoder $g_\phi$; expert sets $\mathcal{E}_{\ell,1}=\{e_{\ell,i}\}_{i=1}^{N}$; buffer $\mathcal{B}_0=\varnothing$; utilization threshold $\eta_{\mathrm{cap}}$
        \FOR{$s=1$ \textbf{to} $S$}
            \STATE Collect demonstrations $\mathcal{D}_s$ from teachers in $\mathcal{G}_s$
            \STATE Build trajectory descriptors and select $\mathcal{R}_s\leftarrow DPP(\mathcal{D}_s)$
            \IF{$s>1$}
                \STATE Evaluate the retained best student $\theta_{s-1}^{*}$ and compute expert-assignment ratios $u_{\ell i}$
                \IF{a layer satisfies $u_{\ell i}\geq\eta_{\mathrm{cap}}$ for all its experts}
                    \STATE Append $\Delta N_{\ell,s}$ new experts to each qualifying layer
                \ENDIF
            \ENDIF
            \STATE Form the offline consolidation set $\widetilde{\mathcal{D}}_s\leftarrow\mathcal{D}_s\cup\mathcal{B}_{s-1}$
            \STATE Train $g_\phi$ independently on observation sequences in $\widetilde{\mathcal{D}}_s$ using InfoNCE
            \STATE Obtain task references $\tau_k^*$ from selected data $\mathcal{B}_{s-1}\cup\mathcal{R}_s$
            \STATE Compute and fix task codes $\mathbf{z}_{\tau_k^*}\leftarrow g_\phi(\tau_k^*)$
            \IF{$s=1$}
                \STATE Train all parameters of the initial student on $\widetilde{\mathcal{D}}_s$
            \ELSE
                \STATE During warm-up: freeze the backbone and old experts; train the router and new experts
                \STATE After warm-up: freeze the backbone and router; train all experts
            \ENDIF
            \STATE Retain the best evaluation checkpoint $\theta_s^{\star}$
            \STATE Update the replay buffer $\mathcal{B}_s\leftarrow\mathcal{B}_{s-1}\cup\mathcal{R}_s$
        \ENDFOR
        \STATE \textbf{return} $\pi_{\theta_S^{\star}}$, $g_{\phi_S}$, $\mathcal{E}_{\ell,S}$, and $\mathcal{B}_S$
    \end{algorithmic}
    \endgroup
\end{algorithm}

\subsection{Teacher--Student Policy Consolidation}
\label{sec:central_model}

\paragraph{Task-specialized teachers and the distillation objective.}
Each teacher $\pi_k^{\mathrm{T}}$ is an independently parameterized continuous-control policy trained by RL for task $k$ according to Equation~\ref{eq:teacher_rl_objective}. CPC does not require teachers to share an architecture or optimizer; each teacher only needs to provide the observation--action demonstrations consumed by the student. In our implementation, PPO actors~\citep{schulman2017proximalpolicyoptimizationalgorithms} generate trajectories $\tau=\{(o_t,a_t,r_t)\}_{t=0}^{H-1}$. At stage $s$, one or more teachers in $\mathcal{G}_s$ contribute distillation dataset $\mathcal{D}_s$, and earlier teachers need not remain accessible after representative trajectories have entered $\mathcal{B}_{s-1}$.

\paragraph{Transformer-MoE central student.}
The central student is implemented by \texttt{SeqTransformerPolicy}. At each action step $t$, a fixed task code $\mathbf{z}_{\tau_k^*}$, computed by the separate context encoder in Section~\ref{sec:replay}, is concatenated with the latest observation, $\mathbf{x}_t=[\mathbf{z}_{\tau_k^*};o_t]$, and projected to the Transformer hidden dimension. The resulting representation receives a learnable positional embedding and passes through stacked Switch Transformer blocks~\citep{fedus2022switch}. Each block retains shared self-attention and replaces its dense feed-forward sublayer with a sparse MoE. The router selects the top-$k$ experts for each representation and combines their outputs using learned routing weights. Two multilayer perceptron (MLP) heads map the final position after layer normalization to the mean and clipped log-standard deviation of the continuous action distribution. Because each expert is a modular feed-forward branch, new experts can be appended without changing the hidden dimension or action interface, enabling the expansion described in Section~\ref{sec:moe}.

\paragraph{Offline teacher-to-student transfer.}
Teachers and the student communicate only through stored trajectories. Each trajectory $\tau$ from task $k$ in $\mathcal{D}_s\cup\mathcal{B}_{s-1}$ is paired with that task's fixed context code $\mathbf{z}_{\tau_k^*}$, and the student regresses the corresponding teacher actions:
\begin{equation}
    \mathcal{L}_{\mathrm{PD}}^{(s)}
    =
    \mathbb{E}_{\tau\sim\mathcal{D}_s\cup\mathcal{B}_{s-1}}
    \left[
    \frac{1}{|\tau|}
    \sum_t
    \left\|
    \mu_\theta(o_t,\mathbf{z}_{\tau_k^*})-a_t
    \right\|_2^2
    \right].
    \label{eq:stage_distillation}
\end{equation}
Thus, RL is confined to teacher-side skill acquisition, whereas student parameters are updated only from offline demonstrations. Environment rollouts are still used to evaluate the student, select checkpoints, and measure expert utilization; they do not supply additional online training data for the student. This distinction separates the cost of acquiring skills from the optimization used to consolidate them. The offline interface supports experience replay for historical supervision, while the MoE structure permits additional capacity when evaluation-time utilization triggers expansion.

Task-specialized teachers and internal MoE experts serve different roles. Teachers are complete policies trained through environment interaction; MoE experts are feed-forward modules inside the central student and learn through distillation. Expanding the student therefore does not insert a teacher policy or require one expert per task. The router can reuse internal experts across teacher behaviors, while new modules provide additional trainable representations within the same action interface. This distinction is central to consolidation: the deployed policy remains a single student rather than an ensemble of retained teachers, and replay carries historical behavioral evidence rather than copies of teacher parameters.

\subsection{Prioritized Trajectory Replay for Stability}
\label{sec:replay}
\label{sec:taskembedding}

Experience replay (ER) preserves old teacher supervision after the corresponding teacher becomes unavailable. CPC implements ER through prioritized trajectory replay (PTR), which selects complete demonstrations for the replay buffer~\citep{liu2023prioritized}. The central student itself uses the latest observation together with a fixed task code at each action step; a complete trajectory is retained because PTR must characterize behavioral coverage and because the context encoder learns from temporally ordered observation histories.

\paragraph{Diversity-prioritized trajectory selection.}
Replay should preserve the range of behaviors demonstrated by each teacher. Selecting only successful or high-return trajectories can repeatedly retain the same dominant solution while discarding alternative state occupancies and action patterns. Under a limited replay budget, CPC instead preserves trajectories that cover complementary modes of a teacher's behavior. Such coverage gives the student supervision over a broader portion of the old policy distribution and provides useful historical targets as the student adapts to later tasks.

Prioritized trajectory replay represents this diversity at the trajectory level. Each observation trajectory is partitioned into non-overlapping temporal slices; the observations within each slice are averaged, and the slice means are concatenated into a descriptor $v_\tau$. Unlike a single trajectory-wide mean, this descriptor retains coarse progression through the rollout; trajectories that visit different regions or reach similar states in different temporal patterns therefore remain distinguishable. With descriptors stacked in $V$, the kernel $L=VV^\top$ measures their similarity. A determinantal point process (DPP) then favors subsets with a large determinant, selecting trajectories that span complementary descriptor directions~\citep{kulesza2012determinantal,chen2023unified}. Algorithm~\ref{alg:cpd} applies this selection to the current demonstrations and appends the resulting subset $\mathcal{R}_s$ to $\mathcal{B}_{s-1}$.

\paragraph{Replay-conditioned context encoder for plasticity.}
The PTR-selected buffer also supports a separate context encoder $g_\phi$, which learns task-discriminative representations from observation sequences rather than individual transitions. At each stage, the encoder is trained independently on observation histories from both current-task trajectories and the historical replay buffer using InfoNCE~\citep{rusak2025infonceidentifyinggaptheory},
\begin{equation}
    \mathcal{L}_{\mathrm{InfoNCE}}
    =-\mathbb{E}_{i}
    \log
    \frac{
    \sum_{j\in\mathcal{P}(i)}
    \exp(\operatorname{sim}(\mathbf{z}_i,\mathbf{z}_j)/\tau_c)
    }{
    \sum_{k\neq i}
    \exp(\operatorname{sim}(\mathbf{z}_i,\mathbf{z}_k)/\tau_c)
    },
    \label{eq:context_encoder}
\end{equation}
where $\mathbf{z}_i$ is the encoding of observation sequence $i$, $\operatorname{sim}$ measures embedding similarity, $\tau_c$ is the temperature, and $\mathcal{P}(i)$ contains other sequences from the same teacher task. After encoder training, the PTR-selected reference observations for task $k$, denoted by $\tau_k^*$ in Algorithm~\ref{alg:cpd}, produce a fixed task embedding $\mathbf{z}_{\tau_k^*}=g_\phi(\tau_k^*)$. This code is held fixed during the stage's student training and evaluation; action prediction requires only the latest observation together with this code. The encoder thus provides a learned alternative to one-hot task representations; the task reference is supplied before evaluation. Its role is to support plasticity through task-discriminative inputs, while replay rehearses older contexts as the encoder learns new ones.

\subsection{Expandable Experts for Plasticity}
\label{sec:moe}

Replay preserves earlier supervision but does not create new trainable capacity. A fixed-capacity central model may struggle to absorb an increasingly diverse task stream. CPC uses expert utilization during evaluation as a heuristic expansion signal: when the existing experts are all substantially engaged, additional experts provide room for subsequent distillation without requiring every new behavior to fit within the same expert bank.

\paragraph{Expert Expansion in MoE.}
After stage $s$, we retain the checkpoint $\theta_s^{\star}$ with the best evaluation performance and record its top-$k$ routing assignments. Because each evaluation input activates $k$ experts, the fraction of assignments received by expert $i$ in layer $\ell$ is
\begin{equation}
    u_{\ell i}
    =
    \frac{1}{k|\mathcal{Q}_s|}
    \sum_{\mathbf{x}\in\mathcal{Q}_s}
    \mathbb{I}\!\left[
    i\in\operatorname{Top-}k(G_\ell(\mathbf{x};\theta_s^{\star}))
    \right],
    \label{eq:expert_utilization}
\end{equation}
where $\mathcal{Q}_s$ denotes evaluation inputs from the selected policy. When every expert in a monitored layer satisfies $u_{\ell i}\geq\eta_{\mathrm{cap}}$, the layer is eligible for expansion. Broad utilization on evaluated tasks indicates that the current expert bank is being actively reused, so CPC adds $\Delta N_{\ell,s}$ new experts to provide trainable capacity for subsequent tasks. This criterion is a practical expansion heuristic based on evaluation-time routing, distinct from the auxiliary loss that encourages balanced routing during training. If some experts remain under-used, CPC first reuses that available capacity.

Expansion does not impose a one-task--one-expert assignment. Each new expert is warm-started from an existing expert with a small perturbation~\citep{chen2016net2netacceleratinglearningknowledge}, preserving a compatible hidden representation. Its initial router bias is negative to reduce the likelihood that an untrained branch dominates routing. During student training, the standard Switch load-balancing term $\mathcal{L}_{\mathrm{bal}}=N_s\sum_{i=1}^{N_s}f_iP_i$, weighted by $\lambda_{\mathrm{aux}}$, is added to the distillation loss; $f_i$ and $P_i$ denote the token fraction and mean routing probability of expert $i$. The context encoder remains optimized separately with the InfoNCE objective in Equation~\ref{eq:context_encoder}. The balancing term discourages routing collapse during training, whereas the assignment ratio in Equation~\ref{eq:expert_utilization} supplies the expansion signal during evaluation.

\paragraph{Two-phase parameter protection for stability.}
Expansion improves plasticity but introduces two sources of interference: updating shared or old-expert parameters can overwrite their functions, and continuing to update the router can redirect old inputs away from the experts that previously served them. CPC therefore protects the expanded student with two parameter masks.

In Phase~1, the shared Transformer backbone and all old experts are frozen, while the router and newly added experts adapt to the incoming demonstrations using the precomputed task codes. The new capacity can specialize and the router can learn where it is useful, while the shared backbone and old-expert parameters remain unchanged. Current and replay trajectories are both present, so routing changes that make old paths inaccessible are penalized through the old action targets.

After the routing warm-up, Phase~2 freezes the shared backbone and router while optimizing all experts on current and replay trajectories. Freezing the router removes routing-parameter updates as one source of interference, but does not guarantee unchanged assignments: updates to earlier experts can alter the representations reaching later routers. Theory for continual MoE likewise motivates terminating gating updates after sufficient training~\citep{li2025theorymixtureofexpertscontinuallearning}. In CPC, the fixed routing parameters and replay supervision play complementary roles: the former constrain routing adaptation, while the latter anchor expert outputs on earlier behaviors. Allowing old and new experts to refine together retains plasticity after warm-up, with historical action targets limiting drift. Parameter protection therefore complements expansion by limiting interference while preserving room for expert adaptation.


\section{Experiments}
\label{sec:experiments}

We conduct the primary evaluation on Meta-World~\citep{mclean2025metaworld} and Continual World~\citep{Wolczyk2021continualworldroboticbenchmark}. The experiments compare CPC with continual-learning baselines, isolate its replay and expandable-expert branches, and visualize the mechanisms behind context encoding and routing. A supplemental experiment is reported separately in Appendix~\ref{appendix:metaworldandlibero}. Our experiments investigate the following questions:

\textbf{Q1}: Can the proposed framework acquire new tasks while retaining previously learned behaviors over a long task sequence?

\textbf{Q2}: Does the framework achieve a favorable balance between plasticity and stability on Meta-World and Continual World?

\textbf{Q3}: How do the two core components---prioritized trajectory replay and expert expansion---complement each other in balancing retention and new-task learning?

\textbf{Q4}: How do the supporting mechanisms---parameter protection, auxiliary routing loss, and the context encoder---support continual student consolidation?

\subsection{Experimental Setup}
\label{sec:experimentSetup}

We summarize the experimental setup below. Task compositions and training configurations are detailed in Appendices~\ref{appendix:benchmarkDetails} and~\ref{appendix:hyperParameters}, respectively.

\paragraph{Benchmarks.}
We conduct experiments mainly on Meta-World and Continual World. Continual World is a continual robot learning benchmark~\citep{Wolczyk2021continualworldroboticbenchmark} that includes two sequential task streams, CW10 and CW20. CW10 contains 10 tasks from Meta-World V1~\citep{yu2020meta}, learned sequentially. CW20 traverses this ten-task sequence twice. The order of tasks is presented in Appendix~\ref{appendix:benchmarkDetails}. We also evaluate continual learning on Meta-World~\citep{mclean2025metaworld}, a robot learning benchmark focusing on multi-task learning (e.g., MT10 and MT25) and meta-learning (e.g., ML45). Our Meta-World experiments use V3 reward functions and a larger task set than CW10/CW20, allowing us to evaluate consolidation across more distinct tasks. Based on Meta-World MT25, we divide the 25 tasks into five groups of five, forming a five-stage continual learning pipeline (Continual MT25). Supplementary experiments are presented in Appendices~\ref{appendix:metaworldandlibero}.

\paragraph{Baselines.}
We compare against four continual policy distillation baselines: \textbf{Sequential fine-tuning}, which sequentially trains the student without an explicit retention mechanism; \textbf{EWC}~\citep{kirkpatrick2017overcoming}, which constrains parameters based on their importance to earlier tasks; \textbf{KL}, which regularizes the current policy toward the preceding checkpoint; and \textbf{TRAC}~\citep{muppidi2024fasttracparameterfreeoptimizer}, which replaces AdamW with a parameter-free optimizer originally proposed for lifelong RL. Additionally, we consider \textbf{Independent}, which allocates a separate model to each stage and serves as an interference-free reference. We evaluate these baselines on both the five-phase Meta-World stream and CW10/CW20. The Continual World results additionally include several published CRL methods.

\paragraph{Ablation Studies.}
The five-phase Meta-World ablation separates the two main branches of CPC. \textbf{ER} retains prioritized replay but removes expert expansion, whereas \textbf{Expert Expansion} removes replay while retaining expert expansion. Further variants remove the context encoder (\textbf{No TE} in Table~\ref{tab:results}), gate freezing, or shared-parameter freezing to assess how these mechanisms support plasticity and stability. The two-phase Meta-World experiment additionally varies the number of retained trajectories and compares two diversity-based selectors under the same buffer budget. Farthest-first sampling (FFS) greedily adds the trajectory descriptor with the largest minimum distance from the selected set, whereas DPP evaluates the diversity of a candidate subset jointly through the determinant of its descriptor-similarity kernel. The routing visualization then examines whether the auxiliary loss reduces expert under-utilization at matched training steps.

\paragraph{Evaluation Metrics.}
We evaluate overall performance, retention, and new-task learning using success rate (SR), backward transfer (BWT), and forward transfer (FWT), respectively; higher values are better. Let $p_k(t)$ be the success rate on task $k$ after stage $t$, $K_t$ the number of tasks seen by stage $t$, and $s(k)$ the stage introducing task $k$. BWT compares performance on earlier tasks with their performance at the end of the stage in which they were introduced ($t>1$): negative values indicate forgetting and positive values indicate improvement. Following~\citet{zheng2025mastering}, task-wise FWT compares the normalized area under the learning curve against a single-task policy trained from scratch. For a $\delta$-step training window, define $\mathrm{AUC}_k=\delta^{-1}\int_{(k-1)\delta}^{k\delta}p_k(u)\,du$ and $\mathrm{AUC}_k^{\mathrm{ref}}=\delta^{-1}\int_0^{\delta}p_k^{\mathrm{ref}}(u)\,du$, where $p_k(u)$ and $p_k^{\mathrm{ref}}(u)$ are the continual and from-scratch success-rate curves. For $\mathrm{AUC}_k^{\mathrm{ref}}<1$, positive FWT indicates faster or better learning relative to this reference. The three metrics are
\begin{equation}
\begin{aligned}
    \mathrm{SR}_t &= \frac{1}{K_t}\sum_{k=1}^{K_t}p_k(t), \\
    \mathrm{BWT}_t &= \frac{1}{K_{t-1}}\sum_{k=1}^{K_{t-1}}\left[p_k(t)-p_k(s(k))\right], \\
    \mathrm{FWT}_k &= \frac{\mathrm{AUC}_k-\mathrm{AUC}_k^{\mathrm{ref}}}{1-\mathrm{AUC}_k^{\mathrm{ref}}}.
\end{aligned}
\label{eq:evaluation_metrics}
\end{equation}

\begin{table*}[t]
\vspace{-3em}
\caption{Continual World results under the CW10 and CW20 protocols. Success rate (SR) is reported on a $[0,1]$ scale; backward transfer (BWT) and forward transfer (FWT) are signed, dimensionless scores, with negative values indicating forgetting and negative transfer, respectively. Independent serves as a reference and is excluded from method rankings. Independent, Finetune, KL, EWC, TRAC, and CPC use a continual policy distillation framework; the remaining methods use continual reinforcement learning (CRL). MT=multi-task, Reg=regularization-based, Struc=structure-based, Reh=rehearsal-based.}
\label{tab:results_cw}
\centering
\resizebox{\textwidth}{!}{
\begin{tabular}{c l|ccc|ccc}
\toprule
\multicolumn{2}{l}{\textbf{Benchmarks-v1}} & \multicolumn{3}{c}{\textbf{CW 10}} & \multicolumn{3}{c}{\textbf{CW 20}} \\
\midrule
\multicolumn{2}{l}{Metrics} & SR & BWT & FWT & SR & BWT & FWT \\
\midrule
\rotatebox[origin=c]{90}{MT} & Independent & 0.934{\scriptsize $\pm$0.011} & -- & -- & 0.933{\scriptsize $\pm$0.011} & -- & -- \\
\midrule
\multirow{4}{*}{\rotatebox[origin=c]{90}{Reg}} & Finetune & 0.123{\scriptsize $\pm$0.007} & -0.696{\scriptsize $\pm$0.126} & -0.068{\scriptsize $\pm$0.126} & 0.183{\scriptsize $\pm$0.013} & -0.555{\scriptsize $\pm$0.029} & 0.128{\scriptsize $\pm$0.006} \\
& KL & 0.141{\scriptsize $\pm$0.005} & -0.719{\scriptsize $\pm$0.078} & -0.095{\scriptsize $\pm$0.052} & 0.205{\scriptsize $\pm$0.004} & -0.647{\scriptsize $\pm$0.075} & -0.094{\scriptsize $\pm$0.059} \\
& EWC & 0.168{\scriptsize $\pm$0.034} & -0.692{\scriptsize $\pm$0.004} & -0.113{\scriptsize $\pm$0.161} & 0.183{\scriptsize $\pm$0.416} & -0.649{\scriptsize $\pm$0.077} & -0.232{\scriptsize $\pm$0.064} \\
& SSDE~\citep{zheng2025mastering} & \textbf{0.950{\scriptsize $\pm$0.021}} & 0.000{\scriptsize $\pm$0.001} & \underline{0.300{\scriptsize $\pm$0.020}} & 0.870{\scriptsize $\pm$0.020} & 0.000{\scriptsize $\pm$0.002} & 0.290{\scriptsize $\pm$0.020} \\
\midrule
\multirow{3}{*}{\rotatebox[origin=c]{90}{Struc}} & TRAC~\citep{muppidi2024fasttracparameterfreeoptimizer} & 0.147{\scriptsize $\pm$0.006} & -0.527{\scriptsize $\pm$0.031} & -0.368{\scriptsize $\pm$0.063} & 0.218{\scriptsize $\pm$0.005} & -0.563{\scriptsize $\pm$0.010} & -0.288{\scriptsize $\pm$0.282} \\
& PackNet~\citep{mallya2018packnet} & 0.830{\scriptsize $\pm$0.040} & \underline{0.000{\scriptsize $\pm$0.000}} & 0.210{\scriptsize $\pm$0.050} & 0.800{\scriptsize $\pm$0.010} & \underline{0.000{\scriptsize $\pm$0.000}} & \underline{0.180{\scriptsize $\pm$0.030}} \\
& CoTASP~\citep{yang2023continual} & 0.730{\scriptsize $\pm$0.110} & \underline{0.000{\scriptsize $\pm$0.000}} & -0.210{\scriptsize $\pm$0.040} & 0.740{\scriptsize $\pm$0.030} & 0.000{\scriptsize $\pm$0.010} & -0.190{\scriptsize $\pm$0.020} \\
\midrule
\multirow{2}{*}{\rotatebox[origin=c]{90}{Reh}} & A-GEM~\citep{chaudhry2018efficient} & 0.140{\scriptsize $\pm$0.050} & -0.730{\scriptsize $\pm$0.010} & 0.280{\scriptsize $\pm$0.010} & 0.070{\scriptsize $\pm$0.020} & -0.700{\scriptsize $\pm$0.010} & 0.130{\scriptsize $\pm$0.030} \\
& ClonEx-SAC~\citep{wolczyk2022disentangling} & 0.860{\scriptsize $\pm$0.020} & -0.020{\scriptsize $\pm$0.020} & \textbf{0.440{\scriptsize $\pm$0.020}} & \underline{0.870{\scriptsize $\pm$0.010}} & -0.020{\scriptsize $\pm$0.010} & \textbf{0.540{\scriptsize $\pm$0.020}} \\
\midrule
& \textbf{CPC(ours)} & \underline{0.908{\scriptsize $\pm$0.011}} & \textbf{0.009{\scriptsize $\pm$0.013}} & 0.144{\scriptsize $\pm$0.061} & \textbf{0.932{\scriptsize $\pm$0.031}} & \textbf{0.002{\scriptsize $\pm$0.050}} & 0.177{\scriptsize $\pm$0.051} \\
\bottomrule
\end{tabular}}
\vspace{-10pt}
\end{table*}

\subsection{Results Analysis}
\label{sec:results}

\subsubsection{Main Results Analysis}
\label{sec:main-results}

\paragraph{Continual World Results.}
Table~\ref{tab:results_cw} reports performance and retention on CW10 and its repeated counterpart, CW20. CPC achieves 90.8\% and 93.2\% success, respectively, with near-zero BWT; its CW20 success rate is comparable to the 93.3\% Independent reference and exceeds those of SSDE and ClonEx-SAC. Repeated exposure may contribute to this improvement, but it does not benefit all methods equally. PackNet falls from 83.0\% to 80.0\%, and SSDE from 95.0\% to 87.0\%, whereas CPC improves while retaining near-zero aggregate BWT. PackNet's near-zero BWT alongside lower final success also illustrates why retention alone is insufficient: preserving old solutions must be accompanied by effective acquisition and refinement of task behaviors. CPC's higher final success and near-zero BWT together support a favorable stability--plasticity balance over the repeated stream.

CPC also retains positive FWT on both streams, although it does not maximize forward transfer. SSDE achieves higher success and FWT on CW10, while ClonEx-SAC achieves higher FWT on both streams. On CW20, however, CPC combines higher final success than either method with near-zero BWT and positive FWT. Together, these results indicate sustained behavioral retention alongside continued learning; the component ablations below examine how replay and expandable capacity contribute to that balance.

\paragraph{Continual MT25 Results.}
Continual MT25 poses a more challenging consolidation problem: its five stages contain heterogeneous V3 tasks, and the task-specific teachers average 81.6\% success (Appendix~\ref{appendix:metaworldandlibero}). CPC reaches $69.4\%$ final success---about 85\% of the teachers' aggregate performance---with BWT of 2.3\%. By contrast, the continual policy distillation baselines finish at 21.4\%--35.9\% with negative BWT, indicating that these distillation approaches leave substantial interference across the heterogeneous stream.

The stage-wise results help interpret the decline in overall success. Success falls from 80.7\% at Stage 2 to 62.3\% at Stage 3, while BWT is $-0.3\%$. Together with the lower teacher success rates for the third task group, this suggests that incoming-task difficulty contributes to the decline, although aggregate BWT alone cannot isolate its contribution. By Stage 5, success rises to 69.4\% with positive BWT, surpassing the 62.8\% Independent reference. Thus, the results support effective cross-stage consolidation with limited aggregate forgetting.

The two evaluations probe complementary aspects of consolidation. Continual MT25 requires the student to absorb heterogeneous teacher groups, whereas CW20 tests whether learning remains effective when previously encountered tasks recur. Success rate and BWT must therefore be interpreted together: high retention can coexist with poor acquisition, and high final success can conceal losses on earlier tasks. CPC's joint results support this balance at the aggregate level; the per-task results in Appendix~\ref{appendix:metaworldandlibero} show where consolidation remains difficult. Moreover, the continual distillation baselines isolate student-side learning under teacher supervision, while the published online-RL results provide a broader performance reference. Their different sources of training data mean that this comparison does not establish equal interaction cost or end-to-end sample efficiency.

\begin{table*}[t]
\vspace{-3em}
\caption{Continual MT25 results on Meta-World. Success rate (SR) and backward transfer (BWT) are reported as percentages; negative BWT indicates forgetting. CPC achieves the highest final success rate (69.4\%) while maintaining positive BWT.}
\label{tab:results}
\centering
\resizebox{\textwidth}{!}{
\begin{tabular}{c l|c|cc|cc|cc|cc}
\toprule
\multirow{2}{*}{} & \multirow{2}{*}{Method} & Stage 1 & \multicolumn{2}{c|}{Stage 2} & \multicolumn{2}{c|}{Stage 3} & \multicolumn{2}{c|}{Stage 4} & \multicolumn{2}{c}{Stage 5} \\
& & SR & SR & BWT & SR & BWT & SR & BWT & SR & BWT \\
\midrule
\multirow{5}{*}{\rotatebox[origin=c]{90}{Baselines}} & EWC & 39.3{\scriptsize $\pm$35.2} & 60.3{\scriptsize $\pm$9.5} & -35.8{\scriptsize $\pm$8.5} & 29.7{\scriptsize $\pm$14.5} & -55.4{\scriptsize $\pm$18.1} & 28.8{\scriptsize $\pm$12.8} & -40.9{\scriptsize $\pm$9.8} & 35.9{\scriptsize $\pm$17.3} & -37.9{\scriptsize $\pm$11.0} \\
& Finetune & \underline{60.0{\scriptsize $\pm$12.0}} & 50.7{\scriptsize $\pm$1.2} & -57.3{\scriptsize $\pm$14.0} & 20.0{\scriptsize $\pm$5.8} & -67.3{\scriptsize $\pm$10.3} & 23.3{\scriptsize $\pm$7.5} & -58.2{\scriptsize $\pm$12.1} & 26.1{\scriptsize $\pm$2.6} & -55.3{\scriptsize $\pm$3.8} \\
& KL & 20.0{\scriptsize $\pm$13.9} & 44.7{\scriptsize $\pm$1.2} & -20.0{\scriptsize $\pm$13.9} & 4.4{\scriptsize $\pm$2.8} & -48.0{\scriptsize $\pm$9.2} & 12.3{\scriptsize $\pm$2.5} & -29.8{\scriptsize $\pm$5.0} & 21.6{\scriptsize $\pm$2.1} & -24.7{\scriptsize $\pm$8.3} \\
& TRAC & 35.0{\scriptsize $\pm$10.5} & 51.0{\scriptsize $\pm$1.2} & -33.0{\scriptsize $\pm$9.5} & 17.7{\scriptsize $\pm$4.1} & -57.5{\scriptsize $\pm$7.7} & 16.2{\scriptsize $\pm$1.7} & -52.3{\scriptsize $\pm$4.3} & 21.4{\scriptsize $\pm$1.8} & -47.8{\scriptsize $\pm$3.1} \\
& Independent & 47.0{\scriptsize $\pm$2.0} & 72.5{\scriptsize $\pm$2.0} & $--$ & 55.0{\scriptsize $\pm$0.9} & $--$ & 58.0{\scriptsize $\pm$2.6} &  $--$ & 62.8{\scriptsize $\pm$2.9} &  $--$ \\
\hline 
\multirow{5}{*}{\rotatebox[origin=c]{90}{Ablations}} & No Freeze-Gating & 54.7{\scriptsize $\pm$3.1} & \underline{74.3{\scriptsize $\pm$7.2}} & 2.7{\scriptsize $\pm$5.0} & \underline{59.3{\scriptsize $\pm$5.3}} & \underline{-1.0{\scriptsize $\pm$7.5}} & \underline{61.2{\scriptsize $\pm$7.8}} & \underline{1.3{\scriptsize $\pm$7.0}} & \underline{64.5{\scriptsize $\pm$1.2}} & 0.8{\scriptsize $\pm$1.4} \\
& No Shared-Frozen & 57.3{\scriptsize $\pm$4.2} & 67.3{\scriptsize $\pm$5.1} & -8.0{\scriptsize $\pm$2.0} & 55.3{\scriptsize $\pm$1.8} & -1.3{\scriptsize $\pm$0.6} & 60.0{\scriptsize $\pm$2.8} & 1.1{\scriptsize $\pm$2.0} & \underline{64.5{\scriptsize $\pm$2.2}} & \textbf{2.8{\scriptsize $\pm$3.3}} \\
& No TE & 39.3{\scriptsize $\pm$5.8} & 68.3{\scriptsize $\pm$6.4} & \textbf{11.3{\scriptsize $\pm$11.7}} & 48.4{\scriptsize $\pm$5.2} & -5.3{\scriptsize $\pm$5.1} & 54.3{\scriptsize $\pm$4.5} & -2.2{\scriptsize $\pm$4.4} & 57.9{\scriptsize $\pm$2.4} & -0.7{\scriptsize $\pm$4.5} \\
& ER & \textbf{61.0{\scriptsize $\pm$8.2}} & 66.5{\scriptsize $\pm$1.9} & -18.0{\scriptsize $\pm$9.5} & 45.6{\scriptsize $\pm$5.6} & -22.0{\scriptsize $\pm$7.5} & 56.2{\scriptsize $\pm$3.2} & -10.4{\scriptsize $\pm$3.3} & 57.2{\scriptsize $\pm$1.4} & -11.0{\scriptsize $\pm$2.8} \\
& Expert Expansion & 50.7{\scriptsize $\pm$2.3} & 52.0{\scriptsize $\pm$2.0} & -44.0{\scriptsize $\pm$4.0} & 16.9{\scriptsize $\pm$3.4} & -67.3{\scriptsize $\pm$4.2} & 24.0{\scriptsize $\pm$1.0} & -55.1{\scriptsize $\pm$3.4} & 28.0{\scriptsize $\pm$3.5} & -51.3{\scriptsize $\pm$4.0} \\
\hline 
& \textbf{CPC(ours)} & 59.3{\scriptsize $\pm$11.0} & \textbf{80.7{\scriptsize $\pm$8.4}} & \underline{6.0{\scriptsize $\pm$8.7}} & \textbf{62.3{\scriptsize $\pm$2.6}} & \textbf{-0.3{\scriptsize $\pm$4.9}} & \textbf{65.3{\scriptsize $\pm$1.8}} & \textbf{2.0{\scriptsize $\pm$2.7}} & \textbf{69.4{\scriptsize $\pm$1.5}} & \underline{2.3{\scriptsize $\pm$2.8}} \\
\bottomrule
\end{tabular}}
\vspace{-10pt}
\end{table*}

\paragraph{Additional results.}
The two-phase Meta-World experiment in Figure~\ref{fig:replay} provides a comparative view of replay-budget and selection effects, while Appendix~\ref{appendix:metaworldandlibero} reports detailed teacher, per-task, and temporal Meta-World results. It also report further evaluates the replay--expansion combination on LIBERO as supplementary cross-benchmark and cross-architecture evidence.

\paragraph{Summary of Q1--Q2.}
Across the two benchmarks, CPC answers Q1 by sustaining one central student through long sequential updates with near-zero or positive BWT. It answers Q2 by combining this retention with strong final success rates on both benchmarks and positive FWT on Continual World: Continual World demonstrates robustness to a repeated long stream, while Meta-World shows that the same framework remains effective when the task groups and teacher quality are substantially more heterogeneous.

\subsubsection{Ablation Study}
\label{sec:ablation}

\begin{figure}[t]
    \vspace{-3em}
    \centering
    \begin{subfigure}[b]{0.56\textwidth}
        \centering
        \adjustbox{width=\linewidth,height=5.2cm,keepaspectratio}{\includegraphics{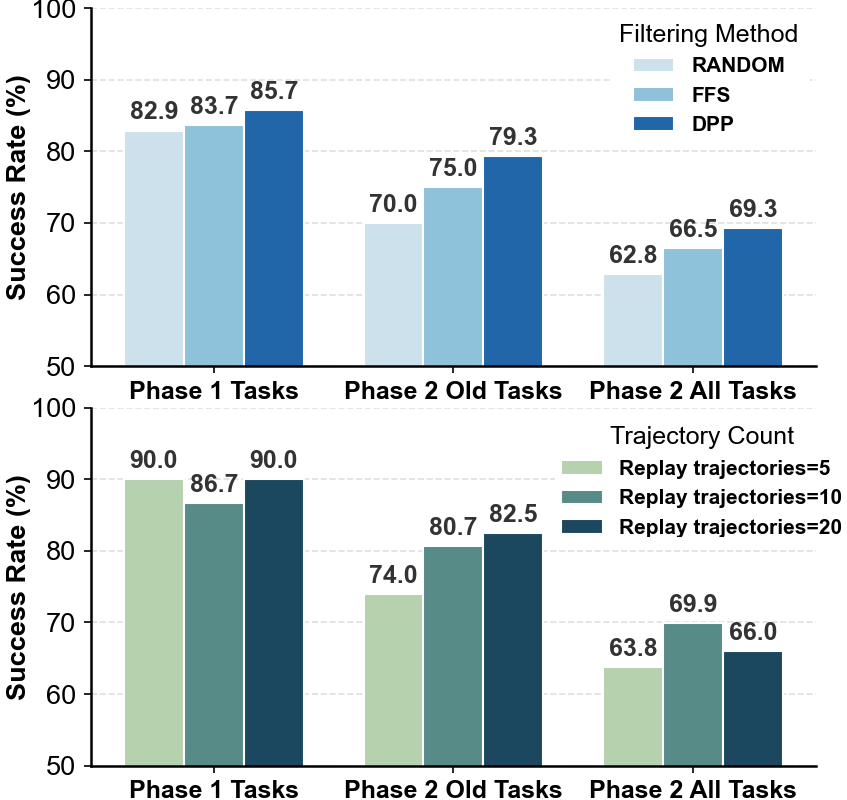}}
        \caption{}
        \label{fig:replay}
    \end{subfigure}
    \hfill
    \begin{subfigure}[b]{0.41\textwidth}
        \centering
        \scriptsize
        \begin{tabular}{@{}c@{\hspace{2pt}}c@{}}
        No aux., L0 & No aux., L3 \\
        \includegraphics[width=0.47\linewidth]{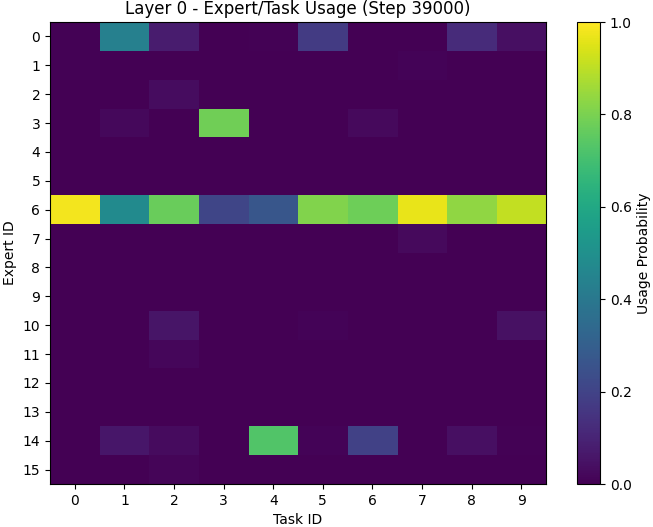} &
        \includegraphics[width=0.47\linewidth]{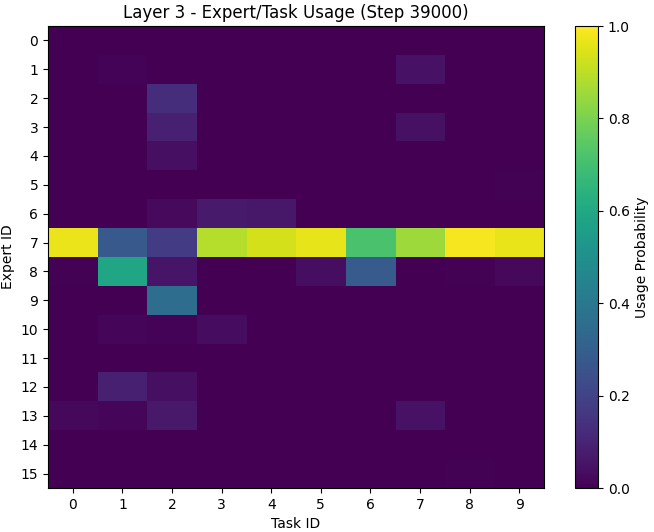} \\
        Use aux., L0 & Use aux., L3 \\
        \includegraphics[width=0.47\linewidth]{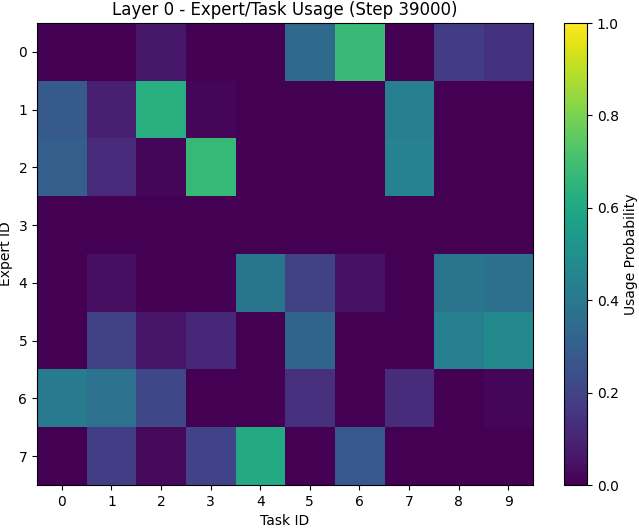} &
        \includegraphics[width=0.47\linewidth]{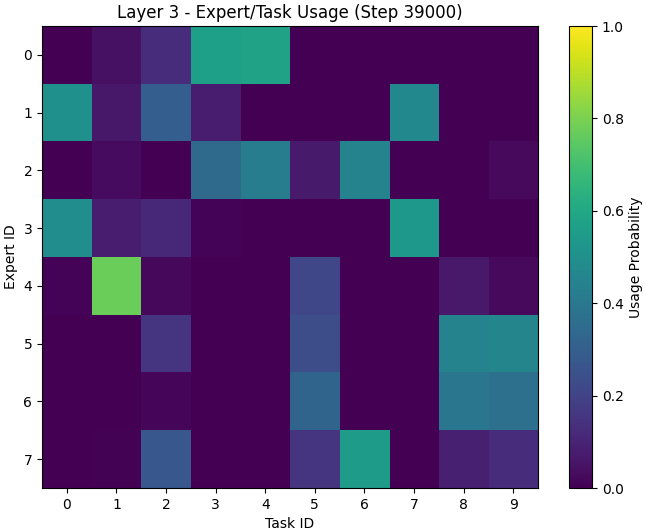}
        \end{tabular}
        \caption{}
        \label{fig:aux_routing_main}
    \end{subfigure}
    \vspace{-5pt}
    \caption{Ablation diagnostics on Meta-World. (a) Retained-task performance under different trajectory budgets and replay-selection rules in the two-phase experiment. (b) Expert assignments at matched training steps without (top) and with (bottom) the auxiliary routing loss; the complete four-layer comparison is in Appendix~\ref{appendix:aux_loss}.}
    \label{fig:ablation_diagnostics}
    \vspace{-10pt}
\end{figure}

\begin{wrapfigure}{r}{0.4\textwidth}
    \vspace{-3em}
    \includegraphics[width=\linewidth]{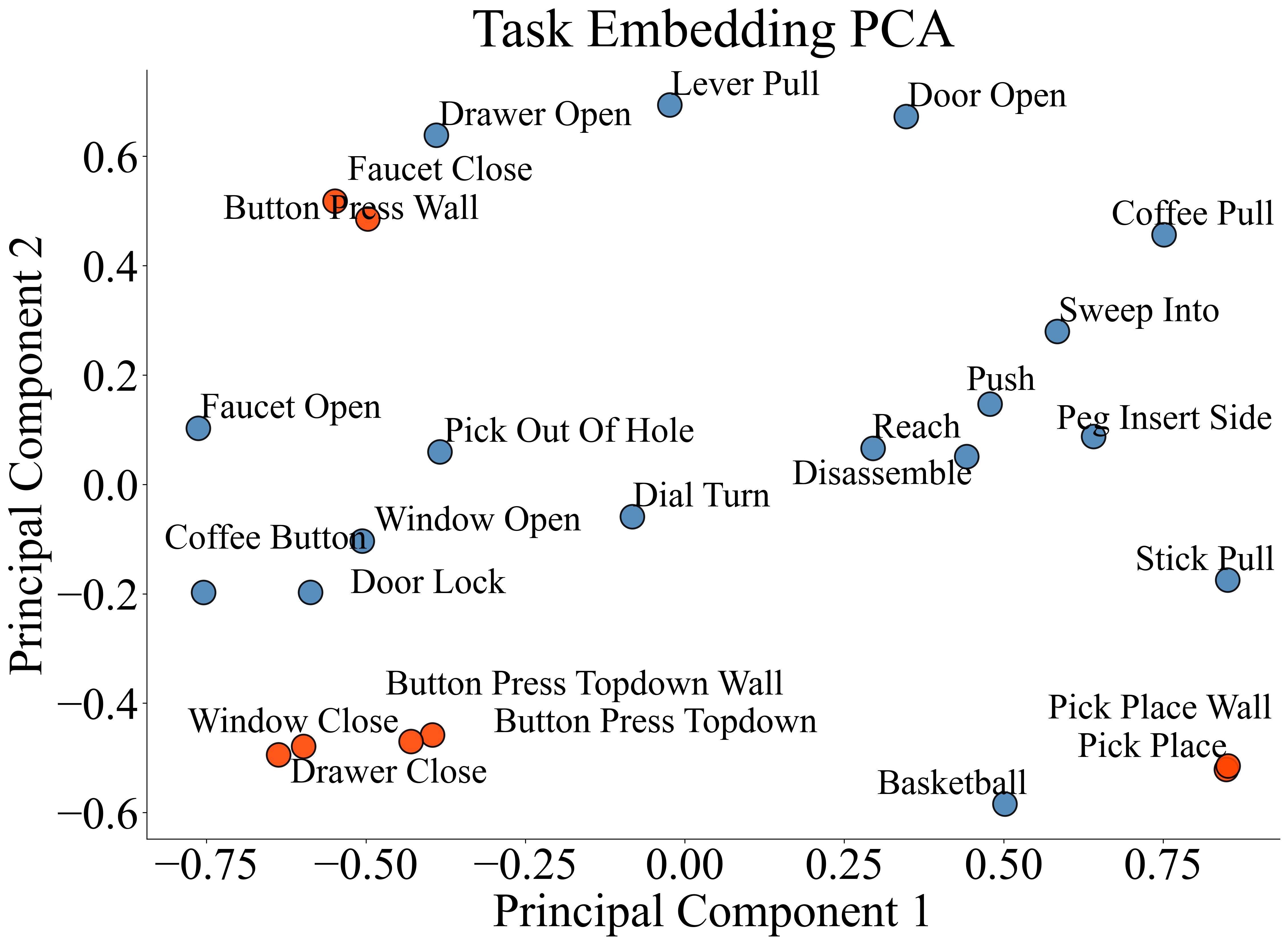}
    \caption{PCA of context representations learned on the Continual MT25.}
    \label{fig:pca}
    \vspace{-3em}
\end{wrapfigure}

The Continual MT25 ablations in Table~\ref{tab:results} address Q3 by isolating replay and expert expansion, and Q4 by testing parameter protection and the context encoder. Figures~\ref{fig:ablation_diagnostics} and~\ref{fig:pca} further examine replay selection, expert utilization, and context representations.

\paragraph{Expandable Experts and Parameter Protection.}
The \textbf{ER} variant retains replay but removes expert expansion. It preserves substantially more behavior than expansion alone, yet finishes at $57.2\%$ success and BWT of $-11.0\%$, consistent with interference remaining under fixed capacity. The full model raises these values to $69.4\%$ and BWT of $2.3\%$, supporting the use of additional capacity alongside historical supervision. The freezing ablations address Q4: removing either gate freezing or shared-parameter freezing lowers final success to $64.5\%$. Their BWT trajectories reveal different effects rather than uniform degradation. Without shared-parameter freezing, Stage 2 BWT is $-8.0\%$, but final BWT recovers to $2.8$; without gate freezing, final BWT is $0.8$. Thus, parameter protection improves the overall learning--retention balance, without necessarily maximizing BWT at every stage.

\paragraph{Prioritized Trajectory Replay.}
The \textbf{Expert Expansion} variant retains expandable experts but removes replay, reducing final success to $28.0\%$ and BWT to $-51.3\%$; in this setting, expandable capacity alone is insufficient to preserve earlier behaviors. Figure~\ref{fig:replay} shows that DPP yields the strongest old-task performance among the tested selectors when applied to the trajectory descriptors. Retaining about ten trajectories from roughly 120 original distillation trajectories already yields strong old-task performance, suggesting that, in this experiment, a DPP-selected buffer of this size provides sufficient behavioral diversity for effective retention.

\paragraph{Auxiliary routing loss.}
Figure~\ref{fig:aux_routing_main} records expert assignments during training at matched steps. Without the auxiliary routing loss, inputs from different tasks concentrate on the same small subset of experts. With the loss enabled, assignments spread across the available experts. The loss discourages early routing collapse, encourages exploration of alternative experts, and exposes them to diverse task signals that can support reusable shared features. Balanced utilization does not require identical expert specialization. These training-time assignments illustrate the intended load-balancing effect; they do not by themselves measure changes in final task performance.

\paragraph{Context encoder.}
Removing the context encoder (\textbf{No TE}) reduces final success rate by 11.5\% and BWT by 3.0\%. The effect is smaller than removing either main branch, which is consistent with the encoder's role as a supporting mechanism rather than the primary source of stability or capacity. Figure~\ref{fig:pca} illustrates the learned context structure: trajectory representations from distinct tasks form separated regions in the projection, while related manipulation behaviors retain local proximity. The fixed, trajectory-derived context gives the student and router a behavior-relevant task representation in place of a one-hot code. The encoder is used in the Continual MT25 implementation, so this evidence concerns its contribution in that setting.

\paragraph{Summary of Q3--Q4.}
For Q3, replay and expert expansion are complementary: historical demonstrations constrain forgetting, while additional capacity improves consolidation beyond replay alone. Neither branch alone matches the full model's final success rate and retention on Continual MT25. The replay-selection study further suggests that a small, diverse trajectory buffer can provide effective historical supervision. 
For Q4, the supporting mechanisms further refine this balance. Removing either freezing component lowers final success, while removing the context encoder reduces both final success and BWT. The routing visualization separately shows that the auxiliary loss broadens expert utilization during training; this observation supports its role in encouraging expert exploration.

\section{Conclusion}

We presented a teacher--student framework for continual policy consolidation in robotic manipulation. The framework decomposes continual robot learning into two stages: training task-specific teachers and sequentially distilling their behaviors into a shared central model. Prioritized trajectory replay supplies historical supervision, while incremental expert expansion provides capacity for new tasks. Parameter protection and task-discriminative context representations support this balance between acquisition and retention. Experiments on Meta-World MT25 and Continual World show strong final performance with low forgetting; aggregate forgetting remains below 5\% in both evaluations. These results demonstrate the potential of the proposed architecture for long-horizon continual learning. Future work will examine whether the framework scales to more diverse task distributions and settings with fewer task-boundary assumptions.

\subsubsection*{Broader Impact Statement}

This work aims to make generalist robot policies more scalable by separating task-specific skill acquisition from continual skill consolidation. Reusing independently trained teachers can reduce repeated training and allow new skills to be integrated while retaining previously learned behaviors. However, the student may inherit teacher errors, and a limited replay buffer may miss rare or safety-critical states; expert expansion also increases model size and computational cost. Our results are limited to simulated manipulation benchmarks, so success rate, BWT, and FWT do not certify real-world reliability. Any deployment should therefore include task-level safety constraints, out-of-distribution monitoring, independent tests after each update, human oversight, and rollback mechanisms. Safety-critical applications require additional domain-specific review and validation beyond the scope of this work.
To facilitate the reproducibility of our results, we have made the source code publicly available (Appendix~\ref{appendix:benchmarkDetails}).



\clearpage\newpage
\bibliography{main}
\bibliographystyle{tmlr}

\clearpage\newpage

\appendix
\section{Benchmark Protocols and Task Composition}
\label{appendix:benchmarkDetails}

We have made the source code publicly available in github~\footnote{https://github.com/Agentic-Intelligence-Lab/Continual-Policy-Distillation/tree/cpc}.

\subsection{Meta-World Task Streams}
Meta-World~\citep{mclean2025metaworld} provides a common Sawyer-arm manipulation interface across heterogeneous tasks. We use the MT10 and MT25 task sets with the V3 reward functions. In the primary five-phase protocol, the 25 tasks are ordered by worker ID and partitioned into five consecutive groups of five tasks, as indicated by the Phase column in Table~\ref{tab:metaworld_results}. The first two groups constitute the ten-task set used in the first phase of the shorter MT10-to-MT25 experiment; the remaining fifteen tasks are introduced together in its second phase.

The Meta-World observation is a 39-dimensional vector describing the end effector, gripper, task-relevant object, goal, and their relative configuration. The central student concatenates this observation with the 16-dimensional task code produced by the context encoder described in Section~\ref{sec:taskembedding}, producing a 55-dimensional input. Actions are four-dimensional, comprising the Cartesian end-effector displacement $(\Delta x,\Delta y,\Delta z)$ and gripper actuation, with each component clipped to $[-1,1]$.

We use V3 dense reward functions for teacher training and the environment-defined binary success condition for evaluation. Episodes have a horizon of 500 steps. During demonstration collection, \verb|terminate_on_success| is disabled so that each stored trajectory covers a complete episode, while object and goal initializations remain randomized. Table~\ref{tab:metaworld_results} reports the task order, teacher success and return, and the corresponding joint-distillation and five-phase CPC success rates. Noted that our CPC results reports here only indicates the task success in current stage, so they are different from those reported in Table~\ref{tab:results} that refer to overall performance. Teacher-training details are given in Appendix~\ref{appendix:ppo}. Returns document teacher performance but should not be compared directly across tasks because Meta-World rewards are task-specific.

\subsection{Continual World Task Stream}
Continual World~\citep{Wolczyk2021continualworldroboticbenchmark} is constructed from Meta-World tasks but standardizes their order for continual reinforcement learning. The original benchmark uses the V1 task definitions, shaped dense rewards during learning, and binary task success for evaluation. Its CW10 sequence is:
\begin{enumerate}
\item \texttt{hammer-v1}
\item \texttt{push-wall-v1}
\item \texttt{faucet-close-v1}
\item \texttt{push-back-v1}
\item \texttt{stick-pull-v1}
\item \texttt{handle-press-side-v1}
\item \texttt{push-v1}
\item \texttt{shelf-place-v1}
\item \texttt{window-close-v1}
\item \texttt{peg-unplug-side-v1}.
\end{enumerate}
CW20 presents this ordered ten-task sequence twice in total. The original Continual World protocol trains an online reinforcement-learning agent for a fixed interaction budget on each task. Our experiment retains the benchmark's V1 task definitions and ordering. Average performance and forgetting are computed from normalized binary success rates, so task-specific dense-reward scales do not enter the reported evaluation metrics. This differs from our Meta-World protocol, which uses V3 task definitions and groups five incoming teachers into each continual-distillation stage.

\subsection{LIBERO Task Suites}
The LIBERO supplementary experiment~\citep{liu2023libero} follows the four-phase order \texttt{LIBERO\_10} $\rightarrow$ \texttt{LIBERO\_GOAL} $\rightarrow$ \texttt{LIBERO\_OBJECT} $\rightarrow$ \texttt{LIBERO\_SPATIAL}. Each suite contains ten manipulation tasks, for a total of 40 task specifications across the stream. Training uses the demonstrations distributed with the corresponding LIBERO suites. The action-model and replay configurations are reported in Appendix~\ref{appendix:libero_hyperparameters}.

\begin{table}[t]
\centering
\caption{Meta-World MT25 task order and per-task results. Phases 1--5 define the five-phase stream, while Phases 1--2 form the initial MT10 group in the two-phase protocol. Teacher results are obtained after at most 10M PPO steps; Joint and CPC report mean success rate $\pm$ standard deviation.}
\label{tab:metaworld_results}
\label{tab:metaworld_results_comparison}
\scriptsize
\setlength{\tabcolsep}{3.5pt}
\resizebox{\textwidth}{!}{%
\begin{tabular}{cclcccc}
\toprule
\textbf{Phase} & \textbf{Worker ID} & \textbf{Task} & \textbf{Teacher success} & \textbf{Teacher return} & \textbf{Joint} & \textbf{CPC} \\
\midrule
\multirow{5}{*}{1} & 0 & reach-v3 & 1.0000 & 140.1236 & 1.00{\scriptsize $\pm$0.00} & 1.00{\scriptsize $\pm$0.00} \\
 & 1 & push-v3 & 0.3200 & 162.2003 & 0.47{\scriptsize $\pm$0.09} & 0.44{\scriptsize $\pm$0.27} \\
 & 2 & pick-place-v3 & 0.8200 & 139.3596 & 0.67{\scriptsize $\pm$0.09} & 0.16{\scriptsize $\pm$0.17} \\
 & 3 & door-open-v3 & 1.0000 & 264.4244 & 1.00{\scriptsize $\pm$0.00} & 0.96{\scriptsize $\pm$0.09} \\
 & 4 & drawer-open-v3 & 1.0000 & 441.4997 & 1.00{\scriptsize $\pm$0.00} & 0.52{\scriptsize $\pm$0.46} \\
\midrule
\multirow{5}{*}{2} & 5 & drawer-close-v3 & 1.0000 & 10.0000 & 1.00{\scriptsize $\pm$0.00} & 1.00{\scriptsize $\pm$0.00} \\
 & 6 & button-press-topdown-v3 & 1.0000 & 122.8903 & 1.00{\scriptsize $\pm$0.00} & 1.00{\scriptsize $\pm$0.00} \\
 & 7 & peg-insert-side-v3 & 0.9800 & 123.7811 & 1.00{\scriptsize $\pm$0.00} & 0.68{\scriptsize $\pm$0.27} \\
 & 8 & window-open-v3 & 1.0000 & 125.6251 & 0.93{\scriptsize $\pm$0.09} & 1.00{\scriptsize $\pm$0.00} \\
 & 9 & window-close-v3 & 1.0000 & 117.3541 & 1.00{\scriptsize $\pm$0.00} & 1.00{\scriptsize $\pm$0.00} \\
\midrule
\multirow{5}{*}{3} & 10 & coffee-pull-v3 & 0.7000 & 58.6285 & 0.40{\scriptsize $\pm$0.16} & 0.60{\scriptsize $\pm$0.31} \\
 & 11 & pick-out-of-hole-v3 & 0.9400 & 69.9379 & 1.00{\scriptsize $\pm$0.00} & 0.76{\scriptsize $\pm$0.26} \\
 & 12 & disassemble-v3 & 0.0000 & 203.7822 & 0.00{\scriptsize $\pm$0.00} & 0.00{\scriptsize $\pm$0.00} \\
 & 13 & pick-place-wall-v3 & 0.4800 & 1786.6363 & 0.40{\scriptsize $\pm$0.16} & 0.08{\scriptsize $\pm$0.10} \\
 & 14 & basketball-v3 & 0.2200 & 121.0556 & 0.00{\scriptsize $\pm$0.00} & 0.00{\scriptsize $\pm$0.00} \\
\midrule
\multirow{5}{*}{4} & 15 & stick-pull-v3 & 0.4800 & 493.3507 & 0.20{\scriptsize $\pm$0.28} & 0.00{\scriptsize $\pm$0.00} \\
 & 16 & button-press-wall-v3 & 0.7400 & 796.8207 & 0.60{\scriptsize $\pm$0.00} & 0.84{\scriptsize $\pm$0.17} \\
 & 17 & faucet-open-v3 & 1.0000 & 216.3948 & 1.00{\scriptsize $\pm$0.00} & 1.00{\scriptsize $\pm$0.00} \\
 & 18 & door-lock-v3 & 1.0000 & 110.0367 & 0.93{\scriptsize $\pm$0.09} & 0.92{\scriptsize $\pm$0.11} \\
 & 19 & lever-pull-v3 & 1.0000 & 282.9568 & 1.00{\scriptsize $\pm$0.00} & 1.00{\scriptsize $\pm$0.00} \\
\midrule
\multirow{5}{*}{5} & 20 & sweep-into-v3 & 0.7400 & 38.9711 & 0.80{\scriptsize $\pm$0.00} & 0.50{\scriptsize $\pm$0.19} \\
 & 21 & faucet-close-v3 & 1.0000 & 192.7584 & 0.73{\scriptsize $\pm$0.25} & 0.88{\scriptsize $\pm$0.11} \\
 & 22 & coffee-button-v3 & 1.0000 & 32.5933 & 1.00{\scriptsize $\pm$0.00} & 1.00{\scriptsize $\pm$0.00} \\
 & 23 & button-press-topdown-wall-v3 & 1.0000 & 152.3181 & 1.00{\scriptsize $\pm$0.00} & 1.00{\scriptsize $\pm$0.00} \\
 & 24 & dial-turn-v3 & 0.9800 & 108.1518 & 1.00{\scriptsize $\pm$0.00} & 0.80{\scriptsize $\pm$0.00} \\
\midrule
\multicolumn{3}{r}{\textbf{Mean}} & \textbf{0.8160} & \textbf{252.4660} & \textbf{0.7652} & \textbf{0.6856} \\
\bottomrule
\end{tabular}}
\end{table}

\clearpage
\section{Training Configurations and Computational Resources}
\label{appendix:hyperParameters}

\subsection{RL Training of Meta-World Teachers}
\label{appendix:ppo}
Task-specialized Meta-World teachers are trained with PPO~\citep{schulman2017proximalpolicyoptimizationalgorithms} using the open-source \texttt{metaworld-algorithms} implementation.\footnote{https://github.com/rainx0r/metaworld-algorithms} Training is capped at 10M environment steps per task. An early-stopping rule terminates a run once its evaluated success rate reaches at least 98\%. Table~\ref{tab:ppo_hyperparameters} lists the complete teacher configuration used for the reported policies.

\begin{table}[ht!]
    \centering
    \caption{Meta-World PPO teacher-training configuration}
    \label{tab:ppo_hyperparameters}
    \begin{tabular}{lrlr}
        \toprule
        \textbf{Parameter} & \textbf{Value} & \textbf{Parameter} & \textbf{Value} \\
        \midrule
        \multicolumn{4}{l}{\textit{PPO configuration}} \\
        \midrule
        Discount factor ($\gamma$) & 0.99 & Baseline type & linear \\
        GAE smoothing ($\lambda$) & 0.97 & Normalize advantages & True \\
        Number of epochs & 20 & Clip value loss & False \\
        Gradient steps & 64 & Target KL & None \\
        \midrule
        \multicolumn{4}{l}{\textit{Policy \& optimization}} \\
        \midrule
        Optimizer & Adam & Max gradient norm & 1.0 \\
        Learning rate & $5e-5$ & Std type & \verb|MLP_Head| \\
        Squash tanh & True & & \\
        \midrule
        \multicolumn{4}{l}{\textit{Training loop}} \\
        \midrule
        Total steps & $5e6$ & Evaluation frequency & $1e5$ \\
        Rollout steps & $20,000$ & & \\
        \bottomrule
    \end{tabular}
\end{table}

\subsection{Meta-World Student Distillation}
Student parameter updates use only collected trajectories, without online interaction to obtain additional training data. Evaluation rollouts are used separately for checkpoint selection and expert-utilization measurement. Table~\ref{tab:central_model_params} records the optimization, Transformer-MoE, and input-context settings used for the Meta-World experiments. Baseline-specific changes are limited to the methods described in Section~\ref{sec:experimentSetup}; the table reports the default CPC configuration.

\begin{table}[ht]
\centering
\caption{Meta-World central-student hyperparameters}
\label{tab:central_model_params}
\setlength{\tabcolsep}{5pt}
\begin{tabular}{@{}ll@{\hspace{1.5em}}ll@{}}
\toprule
\multicolumn{2}{c}{\textit{Optimization} \& \textit{training}} & \multicolumn{2}{c}{\textit{Transformer architecture}} \\
\cmidrule(lr){1-2} \cmidrule(lr){3-4}
\textbf{Hyperparameter} & \textbf{Value} & \textbf{Hyperparameter} & \textbf{Value} \\
\midrule
Optimizer & AdamW & Hidden dimensions & 256 \\
Learning rate & $1e-4$ & Block depth (layers) & 4 \\
Distillation loss & MSE & Number of experts & 8 \\
Distillation epochs & 1000 & MLP multiplier & 4 \\
Distill batch size & 16384 & Causal masking & True \\
Initial $\lambda_{\mathrm{aux}}$ & 0.01 & Use aux loss & True \\
$\lambda_{\mathrm{aux}}$ decay per step & 0.00005 & Layer norm expansion & False \\
Final $\lambda_{\mathrm{aux}}$ & 0.0001 & Use tanh & False \\
Collect steps per expert & 200 $\times$ 500 (env steps) & & \\
\midrule
\multicolumn{4}{c}{\textit{Input} \& \textit{context}} \\
\midrule
Sequence length & 20 & Use task embedding (TE) & True \\
\bottomrule
\end{tabular}
\end{table}

\subsection{Continual World Configuration}
The Continual World experiments use the task stream specified in Appendix~\ref{appendix:benchmarkDetails}. CW20 presents the same ten-task sequence twice, applying the specified number of distillation epochs at each task visit.

The model architecture and training hyperparameters are given in Table~\ref{tab:central_model_params}. CW10/CW20 use the same student architecture, with consolidation performed after each task visit.


\subsection{LIBERO Configuration}
\label{appendix:libero_hyperparameters}
The LIBERO student is implemented in the QwenGR00T framework. Table~\ref{tab:liberosettings} reports its benchmark-specific settings and compares ER with ER+Expert Expansion.

\begin{table}[ht]
\centering
\caption{LIBERO training configurations for ER and ER+Expert Expansion}
\label{tab:liberosettings}
\begin{tabular}{lcc}
\toprule
\textbf{Hyperparameter} & \textbf{ER} & \textbf{Ours} \\
\midrule
\multicolumn{3}{c}{\textit{Optimization} \& \textit{training}} \\
\midrule
max\_train\_steps & 6000 & 6000 \\
num\_warmup\_steps & 1000 & 1000 \\
freeze\_modules & - & before warmup: expert-FFN; \\
&& after: expert-gate \\
enable replay & True & True \\
replay sample ratio & 0.1 & 0.1 \\
replay used ratio & 0.3 & 0.3 \\
evaluation episodes & 50 & 50 \\
\midrule
\multicolumn{3}{c}{\textit{Action model}} \\
\midrule
type & Basic Transformers & Switch Transformers \\
num\_layers & 3 & 3 \\
enable expert expansion & False & True \\
original experts & - & 2 \\
num\_new\_expert (each stage) & - & 2 \\

\bottomrule
\end{tabular}
\end{table}

\subsection{Compute Resources}
\label{app:compute_resource}
Table~\ref{tab:compute_resources} reports the computational resources and measured cost of teacher training, demonstration collection, student distillation, and evaluation. Teacher training uses four RTX 4090 accelerators and requires 103.0M environment steps and 54.5 hours; student distillation uses two accelerators and takes $8.6\pm0.3$ hours without online data collection. Evaluation takes 220 seconds per checkpoint. The host uses an Intel Xeon Platinum 8457C CPU and 503 GB of system memory.

\begin{table}[ht]
\caption{Computational resources and measured training and evaluation costs for the Meta-World MT25 experiments.}
\label{tab:compute_resources}
\resizebox{\textwidth}{!}{%
\begin{tabular}{lllll}
\toprule
\textbf{Pipeline stage} & \textbf{Accelerator} & \textbf{Environment steps} & \textbf{Wall-clock time} & \textbf{Notes} \\
\midrule
PPO teacher training & 4$\times$ RTX 4090 (48GB) & 103.0M actual / 250M cap & 54.5h (all) & Early stopping in 16/25 runs \\
Demonstration collection & 2$\times$ RTX 4090 (48GB) & 0.1M steps/task & 90 s/task & 32 parallel environments \\
Student distillation & 2$\times$ RTX 4090 (48GB) & 0 & $8.6 \pm 0.3$ h & Five random seeds \\
Evaluation & Intel(R) Xeon(R) Platinum 8457C & 50 episodes & 220 s/checkpoint & Measured separately; runs in parallel with training \\
End-to-end pipeline & Mixed accelerators above & --- & approximately 64.5 h & --- \\
\midrule
CPU model & \multicolumn{4}{l}{Intel(R) Xeon(R) Platinum 8457C} \\
System memory & \multicolumn{4}{l}{503 GB} \\
\bottomrule
\end{tabular}}
\end{table}

\clearpage
\section{Detailed Meta-World Results}
\label{appendix:metaworldandlibero}

\subsection{Teacher Performance and Per-Task Outcomes}
The MT10 and MT25 teacher collections achieve average success rates of 91.2\% and 81.6\%, respectively. Their task order, success rates, and task-specific returns are consolidated in Table~\ref{tab:metaworld_results}. Sixteen of the 25 MT25 teachers reach the early-stopping criterion before the 10M-step cap, explaining the reduction in actual teacher-training interaction reported in Table~\ref{tab:compute_resources}.

The per-task results reveal two distinct sources of downstream error. First, several tasks are already difficult for the PPO teachers: \texttt{disassemble-v3} has zero teacher success, while \texttt{basketball-v3}, \texttt{push-v3}, \texttt{pick-place-wall-v3}, and \texttt{stick-pull-v3} obtain success rates between 0.22 and 0.48. The zero joint-distillation and CPC scores on \texttt{disassemble-v3} and \texttt{basketball-v3} are therefore consistent with limited successful teacher demonstrations rather than continual forgetting alone. Second, some tasks with strong teachers remain difficult to consolidate. For example, \texttt{pick-place-v3}, \texttt{drawer-open-v3}, and \texttt{peg-insert-side-v3} have teacher success rates of 0.82, 1.00, and 0.98, but their CPC success rates are 0.16, 0.52, and 0.68, respectively. These gaps identify skill-integration difficulties that teacher quality alone does not explain.

Conversely, the student occasionally achieves a higher success rate than its task-specific teacher: CPC reaches 0.44 on \texttt{push-v3} and 0.84 on \texttt{button-press-wall-v3}, compared with teacher success rates of 0.32 and 0.74. These observations are consistent with cross-task reuse, but a controlled transfer study would be needed to distinguish this explanation from other sources of variation. Because dense reward scales differ by task, the return column is retained only to document teacher training and is not used to rank task difficulty across the benchmark.

\subsection{Distillation Results}
Joint distillation provides a full-data reference setting in which the student is optimized with all task demonstrations available together. Table~\ref{tab:multitasklearning} compares the resulting MT10 student with the average teacher reference and published multi-task methods. The central student obtains $88.9 \pm 1.9$\% success, recovering approximately 97.5\% of the $91.2 \pm 1.4$\% teacher reference. Its mean success rate is also close to the reported MOORE result of $88.7 \pm 5.6$\%. However, the methods use different training allocations---10M steps per teacher plus 3M distillation steps for our framework versus 20M steps for the listed multi-task baselines---so the table should be interpreted as a contextual performance comparison rather than a strict compute-matched ranking.

\begin{table}[ht!]
    \vspace{0pt}
    \centering
    \caption{Contextual MT10 comparison. Our framework uses 60k collection steps from each teacher and 3M distillation steps, whereas the reported multi-task baselines use 20M training steps.}
    \begin{tabular}{l|c}
    \toprule
       \textbf{Methods} & \textbf{Success rate (\%)} \\
    \midrule
       SAC \citep{yu2020meta}  & 61.9 {\scriptsize $\pm$ 3.3} \\
       MTSAC \citep{yu2020meta}  & 62.9 {\scriptsize $\pm$ 8.0} \\
       Soft-Module \citep{yang2020multi}  & 63.0 {\scriptsize $\pm$ 4.2} \\
       CARE \citep{sodhani2021multi}  & 76.0 {\scriptsize $\pm$ 6.9} \\
       PaCo \citep{sun2022paco} & 85.4 {\scriptsize $\pm$ 4.5} \\
       MOORE \citep{hendawy2023multi} & \underline{88.7 {\scriptsize $\pm$ 5.6}} \\
    \midrule
       AvgTeachers (ours) & 91.2 {\scriptsize $\pm$ 1.4} \\
       \textbf{Central model (ours)} & \textbf{88.9 {\scriptsize $\pm$ 1.9}} \\
    \bottomrule
    \end{tabular}
    \label{tab:multitasklearning}
    \vspace{-10pt}
\end{table}

Table~\ref{tab:metaworld_results} provides the corresponding task-level view. Joint distillation matches the success rates of several high-performing teachers, including \texttt{reach-v3}, \texttt{door-open-v3}, \texttt{drawer-open-v3}, and multiple button, faucet, and lever tasks. Several of its largest failures occur on tasks with low teacher success, but residual gaps on tasks such as \texttt{coffee-pull-v3} and \texttt{stick-pull-v3} also show that simultaneous access to all demonstrations does not guarantee faithful recovery of every teacher behavior.

\subsection{Five-Phase Temporal Behavior}
\begin{figure*}[h!]
    \centering
    \begin{subfigure}[b]{0.44\textwidth}
        \centering
        \adjustbox{width=\linewidth,height=6cm,keepaspectratio}{\includegraphics{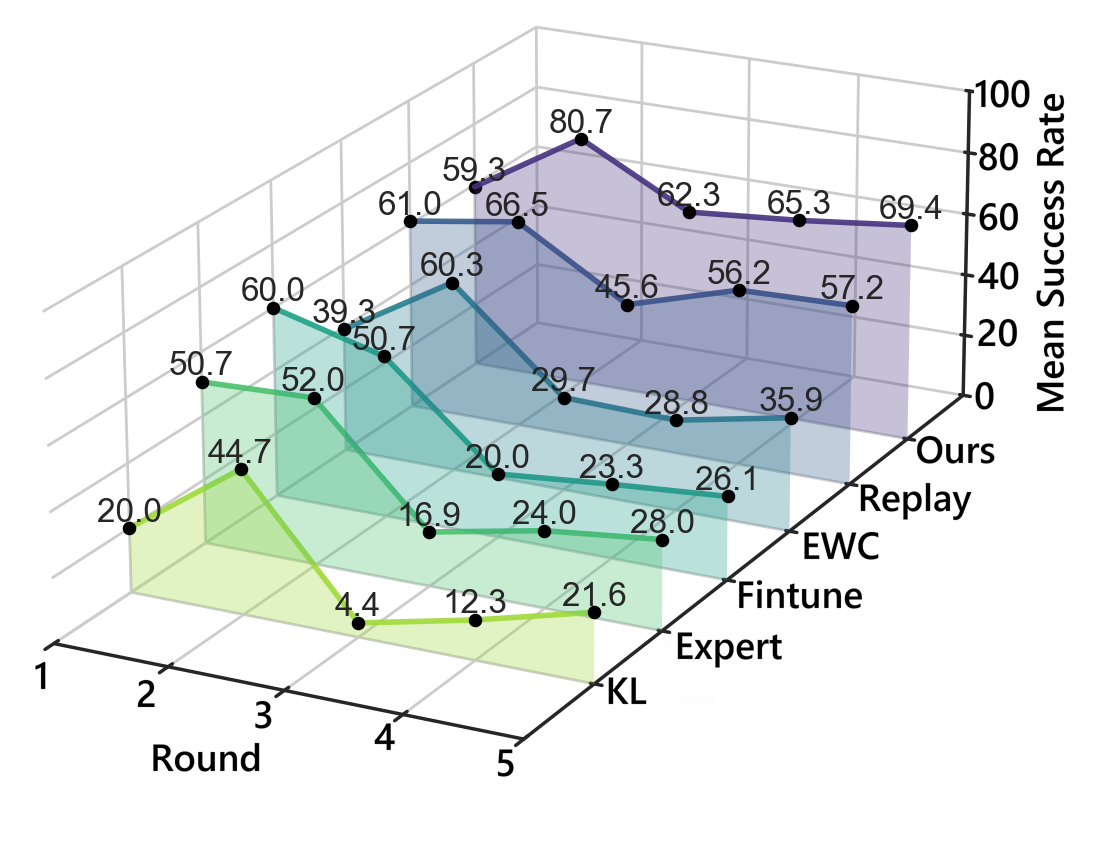}}
        \caption{}
        \label{fig:allScore}
    \end{subfigure}
    \hfill
    \begin{subfigure}[b]{0.54\textwidth}
        \centering
        \adjustbox{width=\linewidth,height=6cm,keepaspectratio}{\includegraphics{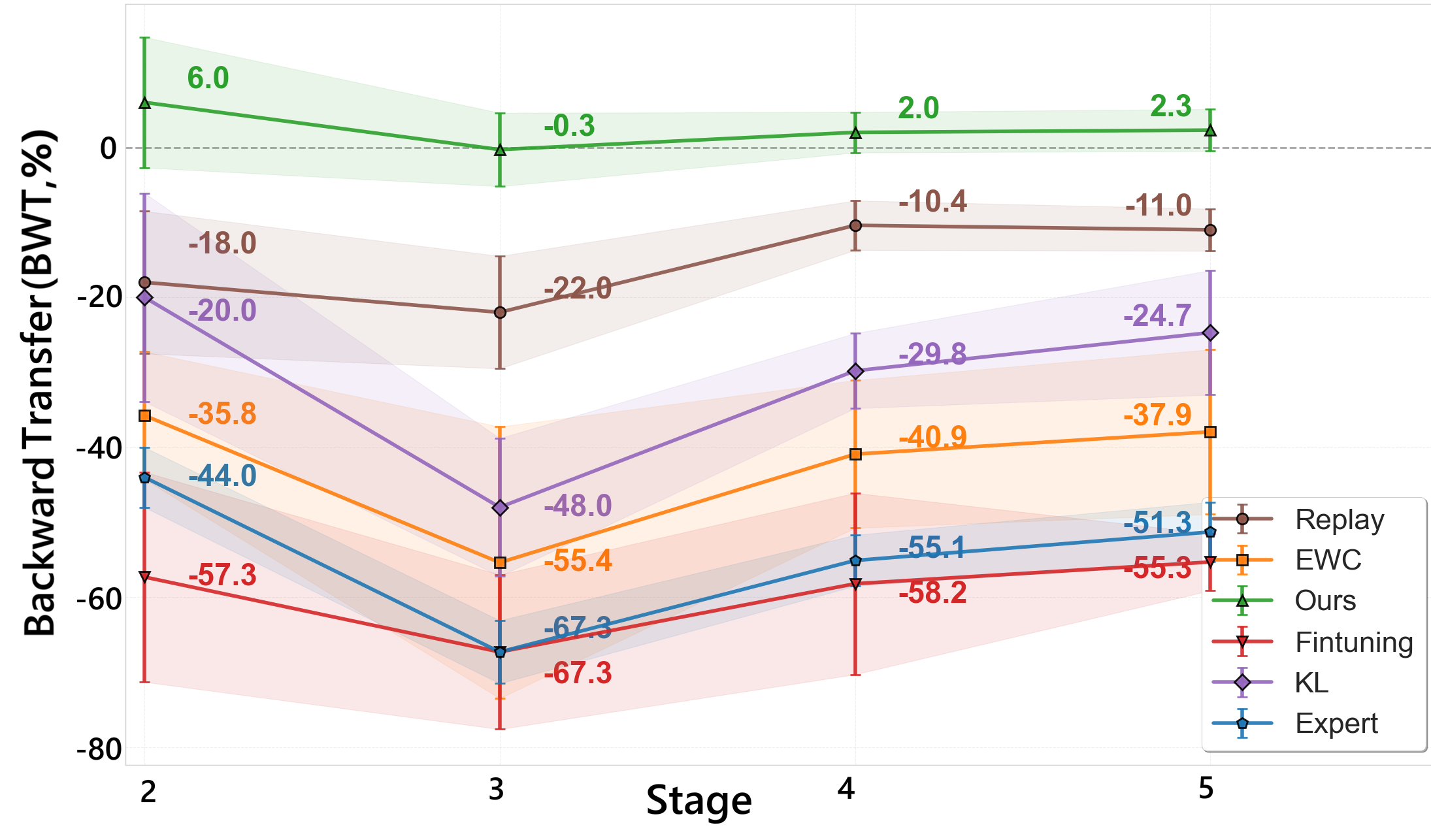}}
        \caption{}
        \label{fig:BWT}
    \end{subfigure}
    \vspace{-5pt}
    \caption{Supplementary curves for the Meta-World MT25 five-phase experiment. (a) Stage-wise success rate and (b) backward transfer. These curves provide the temporal view of the aggregate results reported in Table~\ref{tab:results}.}
    \label{fig:result_analysis}
    \vspace{-10pt}
\end{figure*}

Figure~\ref{fig:result_analysis} complements the stage summaries in Table~\ref{tab:results} by showing when performance changes occur. The overall-performance curves show the widening gap between CPC and the baselines as additional task groups arrive. The BWT curve provides the corresponding retention view: baseline degradation accumulates across stages, whereas CPC remains near zero or positive in the later phases. The figure provides temporal evidence for the main conclusion, while Table~\ref{tab:results} reports the exact stage-wise values.

\subsection{LIBERO Supplementary Results}

\begin{figure*}[h!]
    \centering
    \adjustbox{width=\linewidth,height=5cm,keepaspectratio}{\includegraphics{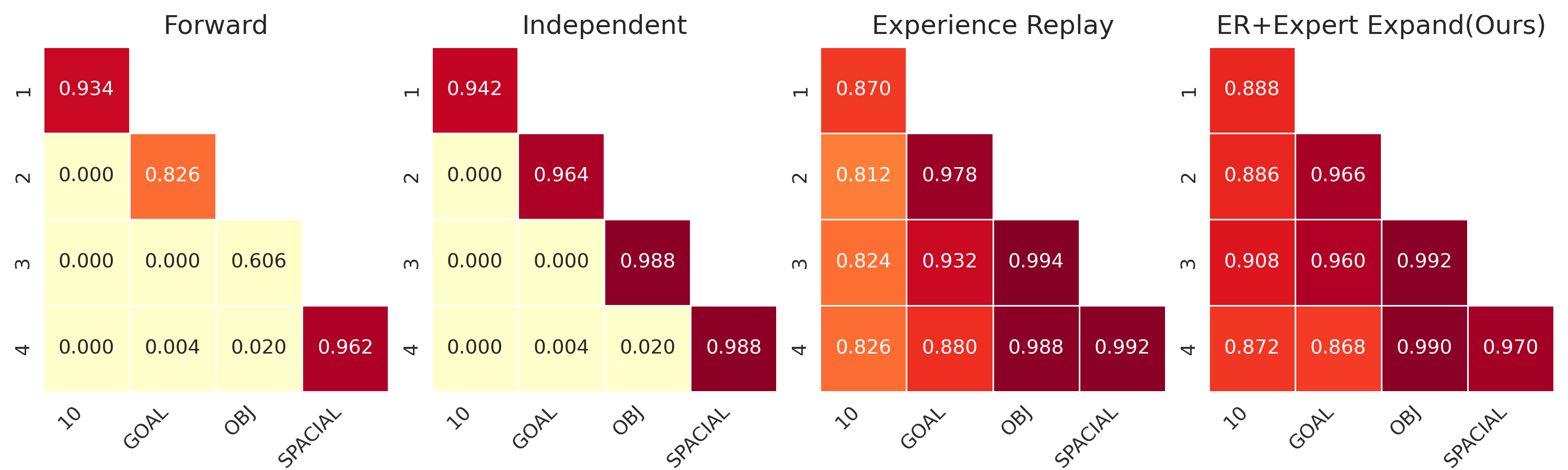}}
    \vspace{-5pt}
    \caption{Success-rate matrices showing average performance on each task suite after each stage of the four-phase LIBERO sequence.}
    \label{fig:libero_baselines}
    \vspace{-10pt}
\end{figure*}

We evaluate the replay--expansion combination on the four-phase sequence \texttt{LIBERO\_10} $\rightarrow$ \texttt{LIBERO\_GOAL} $\rightarrow$ \texttt{LIBERO\_OBJECT} $\rightarrow$ \texttt{LIBERO\_SPATIAL} using the \verb|QwenGR00T| vision--language--action (VLA) model. Figure~\ref{fig:libero_baselines} reports the average success rate after each stage. ER+Expert Expansion reaches a final average success rate of \textbf{92.5\%}, compared with \textbf{92.1\%} for ER alone. The $0.4$-percentage-point difference is a small observed gain from combining expansion with replay in this architecture. Statistical significance has not been established, and the short task stream limits conclusions about long-horizon learning.

\clearpage
\section{Further Analysis of the Auxiliary Routing Loss}
\label{appendix:aux_loss}

Table~\ref{tab:gating_vis} compares expert assignments at matched training steps. Without the auxiliary routing objective, the router concentrates a large fraction of inputs on only one or two experts; this imbalance is most evident in the first and fourth MoE layers. Enabling the auxiliary loss distributes assignments more evenly across the available experts in all four layers, indicating that the objective mitigates routing collapse and reduces persistent expert under-utilization.

This comparison illustrates how the auxiliary term changes routing behavior during training. The broader expert utilization supports the load-balancing interpretation in the main ablation analysis; final success and forgetting must be assessed separately.

\begin{table}[ht]
\centering
\caption{Expert-routing visualizations at matched training steps, without (top) and with (bottom) the auxiliary routing loss. Each column corresponds to one MoE layer.}
\label{tab:gating_vis}

\begin{tabular}{@{}c@{\hspace{2pt}}c@{\hspace{2pt}}c@{\hspace{2pt}}c@{}}
\includegraphics[width=0.24\linewidth]{figures/no_aux_loss_0.png} &
\includegraphics[width=0.24\linewidth]{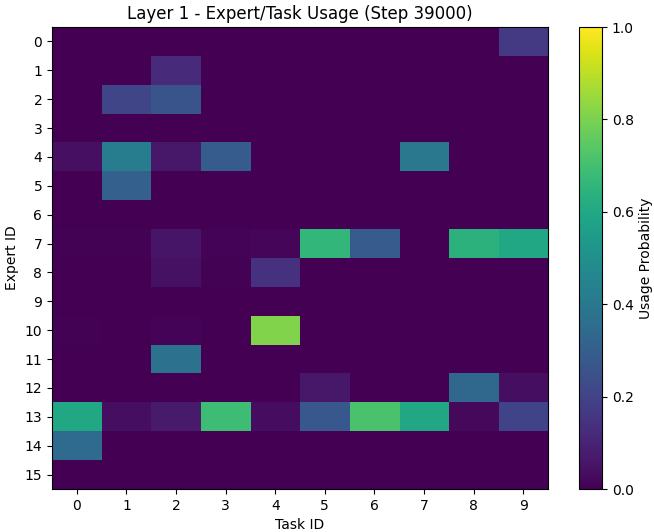} &
\includegraphics[width=0.24\linewidth]{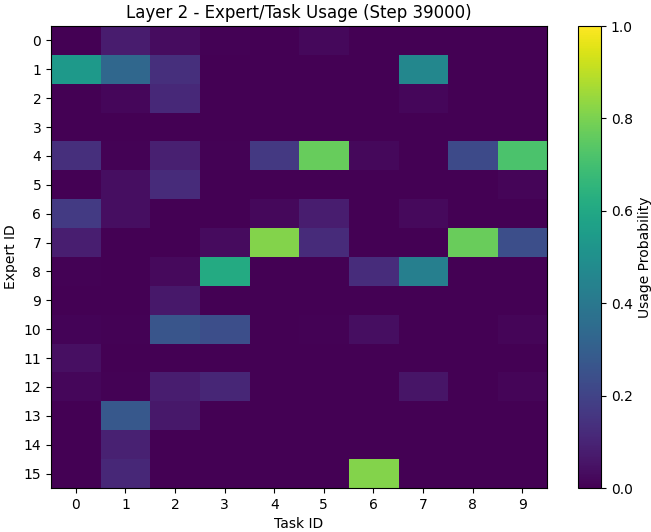} &
\includegraphics[width=0.24\linewidth]{figures/no_aux_loss_3.png}
\end{tabular}
\textbf{$\Downarrow$ Enable auxiliary loss $\Downarrow$} \\
\vspace{0.3em}
\begin{tabular}{@{}c@{\hspace{2pt}}c@{\hspace{2pt}}c@{\hspace{2pt}}c@{}}
\includegraphics[width=0.24\linewidth]{figures/with_aux_0.png} &
\includegraphics[width=0.24\linewidth]{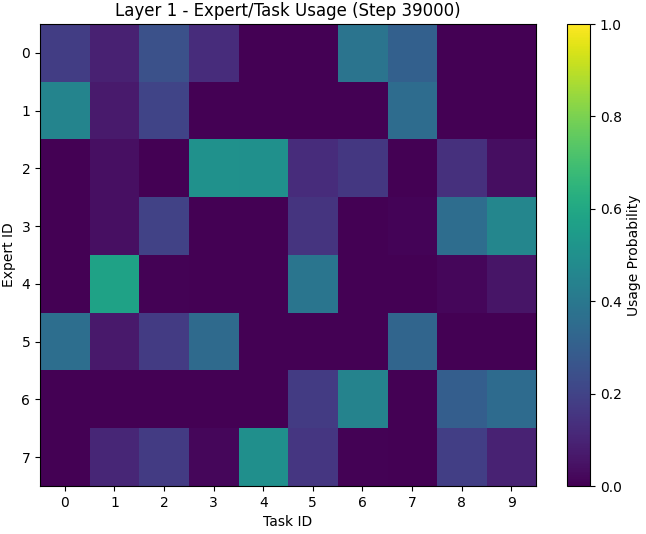} &
\includegraphics[width=0.24\linewidth]{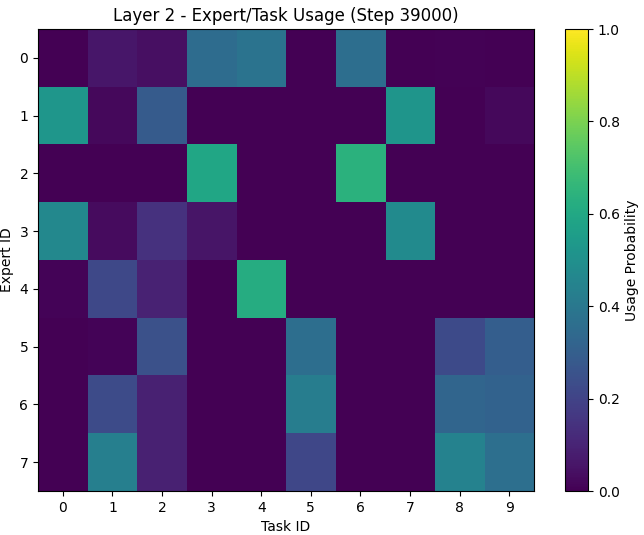} &
\includegraphics[width=0.24\linewidth]{figures/with_aux_3.png} \\
Layer\_0 & Layer\_1 & Layer\_2 & Layer\_3
\end{tabular}

\end{table}

\end{document}

%% file: math_commands.tex
\usepackage{amsmath,amsfonts,bm}

\def\eqref#1{equation~\ref{#1}}

\def\1{\bm{1}}

\DeclareMathAlphabet{\mathsfit}{\encodingdefault}{\sfdefault}{m}{sl}
\SetMathAlphabet{\mathsfit}{bold}{\encodingdefault}{\sfdefault}{bx}{n}

